\documentclass[times,twocolumn,final,authoryear]{elsarticle}

\usepackage{ycviu}
\usepackage{framed,multirow}
\usepackage{hyperref}
\hypersetup{
    colorlinks=true,
    linkcolor=blue,
    filecolor=magenta,      
    urlcolor=cyan,
}
\usepackage{algorithm}
\usepackage[noend]{algpseudocode}
\usepackage{graphicx, caption, subcaption}

\usepackage{amssymb}
\usepackage{amsmath}
\usepackage{mathtools}
\usepackage{latexsym}
\usepackage[toc,page]{appendix}

\usepackage{url}
\usepackage{xcolor}
\definecolor{newcolor}{rgb}{.8,.349,.1}
\newcommand\red[1]{{\color{black}#1}}

\journal{Computer Vision and Image Understanding}

\begin{document}

\clearpage
\thispagestyle{empty}

\ifpreprint
  \vspace*{-1pc}
\else
\fi

\ifpreprint
  \setcounter{page}{1}
\else
  \setcounter{page}{1}
\fi

\begin{frontmatter}

\title{Simultaneous Compression and Quantization:\\A Joint Approach for Efficient Unsupervised Hashing}

\author[1]{Tuan \snm{Hoang}\corref{cor1}} 
\cortext[cor1]{Corresponding author:}
\ead{nguyenanhtuan_hoang@mymail.sutd.edu.sg}
\author[2]{Thanh-Toan \snm{Do}}
\author[3]{Huu \snm{Le}}
\author[1]{Dang-Khoa \snm{Le-Tan}}
\author[1]{Ngai-Man \snm{Cheung}}

\address[1]{Singapore University of Technology and Design, Singapore}
\address[2]{University of Liverpool, Liverpool, UK}
\address[3]{Chalmers University of Technology, Sweden}


\begin{abstract}
For unsupervised data-dependent hashing, the two most important requirements are to preserve similarity in the low-dimensional feature space and to minimize the binary quantization loss.
A well-established hashing approach is Iterative Quantization (ITQ), which addresses these two requirements in separate steps.
In this paper, we revisit the ITQ approach and propose novel formulations and algorithms to the problem.
Specifically, we 
propose a novel approach, named \textbf{Simultaneous Compression and Quantization (SCQ)}, to jointly learn to compress (reduce dimensionality) and binarize input data in a single formulation under strict orthogonal constraint. 
With this approach, we introduce a loss function and its relaxed version,
termed {Orthonormal Encoder (OnE)} and {Orthogonal Encoder (OgE)} respectively, which involve  challenging binary and orthogonal constraints. 
We propose to attack the optimization using novel algorithms based on recent advance in cyclic coordinate descent approach. 
Comprehensive experiments on unsupervised image retrieval demonstrate  that our proposed methods consistently outperform other state-of-the-art hashing methods.
Notably, our proposed methods outperform recent deep neural networks and GAN based hashing in accuracy, while being very computationally-efficient.

%
\end{abstract}


\end{frontmatter}


\newcommand{\Sc}{s}
\newcommand{\B}{\mathbf{B}}
\newcommand{\V}{\mathbf{V}}
\newcommand{\X}{\mathbf{X}}
\newcommand{\W}{\mathbf{W}}
\newcommand{\R}{\mathbf{R}}
\newcommand{\I}{\mathbf{I}}
\newcommand{\1}{\mathbf{\underline{1}}}
\newcommand{\0}{\mathbf{\underline{0}}}
\newcommand{\hX}{\mathbf{\hat{X}}}
\newcommand{\tr}{\textrm{tr}}

\def\x{{\mathbf x}}
\def\y{{\mathbf y}}
\def\w{{\mathbf w}}
\def\h{{\mathbf h}}
\def\cc{{\mathbf c}}
\def\I{{\mathbf 1}}
\def\q{{\mathbf q}}

\def\U{{\mathbf U}}
\def\B{{\mathbf B}}
\def\X{{\mathbf X}}
\def\Y{{\mathbf Y}}
\def\R{{\mathbb R}}
\def\H{{\mathbf H}}
\def\Z{{\mathbf Z}}
\def\W{{\mathbf W}}
\def\V{{\mathbf V}}
\def\Q{{\mathbf Q}}
\def\I{{\mathbf I}}
\def\1{{\mathbf 1}}
\def\S{{\mathbf S}}
\def\V{{\mathbf V}}
\def \reducevspace {-0.7em}


\newcommand{\ba}{\mathbf{a}}
\newcommand{\bb}{\mathbf{b}}
\newcommand{\bc}{\mathbf{c}}
\newcommand{\bk}{\mathbf{k}}
\newcommand{\bo}{\mathbf{o}}
\newcommand{\bs}{\mathbf{s}}
\newcommand{\bt}{\mathbf{t}}
\newcommand{\bu}{\mathbf{u}}
\newcommand{\bv}{\mathbf{v}}
\newcommand{\bw}{\mathbf{w}}
\newcommand{\bx}{\mathbf{x}}
\newcommand{\by}{\mathbf{y}}
\newcommand{\bz}{\mathbf{z}}

\newcommand{\bA}{\mathbf{A}}
\newcommand{\bB}{\mathbf{B}}
\newcommand{\bC}{\mathbf{C}}
\newcommand{\bD}{\mathbf{D}}
\newcommand{\bE}{\mathbf{E}}
\newcommand{\bF}{\mathbf{F}}
\newcommand{\bG}{\mathbf{G}}
\newcommand{\bI}{\mathbf{I}}
\newcommand{\bL}{\mathbf{L}}
\newcommand{\bO}{\mathbf{O}}
\newcommand{\bR}{\mathbf{R}}
\newcommand{\bS}{\mathbf{S}}
\newcommand{\bT}{\mathbf{T}}
\newcommand{\bU}{\mathbf{U}}
\newcommand{\bV}{\mathbf{V}}
\newcommand{\bW}{\mathbf{W}}
\newcommand{\bX}{\mathbf{X}}
\newcommand{\bY}{\mathbf{Y}}
\newcommand{\bZ}{\mathbf{Z}}

\newcommand{\sign}{\mathrm{sign}}
\section{Introduction}
For decades, image hashing has  been an active research field in vision community (\cite{LSH,ITQ, SH,Self-taught_Hash}) due to its  advantages in storage and computation speed for similarity search/retrieval under specific conditions (\cite{ITQ}). Firstly, the binary code should be short so as to the whole hash table can fit in the memory. Secondly, the binary code should preserve the similarity, i.e., (dis)similar images have (dis)similar hashing codes in the Hamming distance space. Finally, the algorithm to learn parameters should be fast and for unseen samples, the hashing method should produce the hash codes efficiently. It is very challenging to simultaneously satisfy
all three requirements, especially, under the binary constraint which leads to an NP-hard mixed-integer optimization problem. In this paper, we aim to tackle all these challenging conditions and constraints. 

The proposed hashing methods in literature can be categorized into data-independence (\cite{SimSearchHighDim,KLSH,LSB}) and data-dependence; in which, the latter recently receives more attention in both (semi-)supervised (\cite{SDP_AL,learn2hash_binaryreconst,FSH,SH_kernel,hamming_learning,SDH,Chen_2018_CVPR,Cao_2018_CVPR,8237358,7780596,7301269,7298947,}) and unsupervised (\cite{BA,UH-BDNN,DBLP:conf/cvpr/DoTPC17,SAH-TIP19,ITQ,KMH,SpH,SADS,ARE,8057811,8320817,8031367,8099999,8296917,8801918}) manners. Supervised hashing have shown superior performance over unsupervised hashing. However, in practice, labeled datasets are limited and costly;
hence, in this work, we focus only on the unsupervised setting. We refer readers to recent surveys (\cite{learnhashforlargescale, surveylearntohash, surveyhashforsimilarity, survey_learn2hash}) for more detailed reviews of data-independent/dependent hashing methods.



\subsection{Related works}
\red{
The most relevant work to our proposal is Iterative Quantization (ITQ) (\cite{ITQ}), which is a very fast and competitive hashing method. 
The fundamental of ITQ is two folds. Firstly, to achieve low-dimensional features, it uses the well-known Principle Component Analysis (PCA) method. PCA maximizes the variance of projected data and keeps dimensions pairwise uncorrelated. Hence, the low-dimension data, projected using the top PCA component vectors, can preserve data similarity well. Secondly, minimizing the binary quantization loss using an orthogonal rotation matrix strictly maintains the data pairwise distance. As a result, ITQ learns binary codes that can highly preserve the local structure of the data. However, optimizing these two steps separately, especially when no binary constraint is enforced in the first step, i.e., PCA, leads to suboptimal solutions. In contrast, we propose to jointly optimize the projection variance and the quantization loss.

Other works that are highly relevant to our proposed method are Binary Autoencoder (BA) (\cite{BA}), UH-BDNN (\cite{UH-BDNN}), DBD-MQ (\cite{8099999}), and Stacked convolutional AutoEncoders (SAE) (\cite{8296917}).
In these methods, the authors proposed to combine the data dimension reduction and binary quantization into a single step by using encoder of autoencoder, while the decoder encourages (dis)similar inputs map to (dis)similar binary codes. However, the reconstruction criterion is not a direct way for preserving the similarity (\cite{UH-BDNN}). 
Additionally, although achieving very competitive performances, UH-BDNN and DBD-MQ are based on the deep neural network (DNN); hence, it is difficult to produce the binary code computationally-efficiently. Particularly, given an extracted CNN feature, these methods require a forward propagation through multiple fully-connected and activation layers to produce the binary code. While our proposed method only requires a single linear transformation, i.e., one BLAS operation ($\texttt{gemv}$ or $\texttt{gemm}$), and a comparison operation.
}

\smallskip

Recently, many works (\cite{DeepBit,8099999,BGAN}) leverage the powerful capability of Convolution Neuron Network (CNN) to jointly learn the image representations and binary codes. 
However, due to the non-smooth property of the binary constraint causing the ill-gradient in back-propagation, these methods resort to relaxation or approximation. As a result, even thought achieving high-discriminative image representations, these methods can only produce sub-optimal binary codes. 
In the paper, we show that 
by directly considering the binary constraint, our methods can obtain much better binary codes. Hence, higher retrieval performances can be achieved. This emphasizes the necessity of having an effective method to preserve the discrimination power of high-dimensional CNN features in very compact binary representations, i.e., effectively handling the challenging binary and orthogonal constraints. 

\red{
Besides, several works has been proposed to handle the difficulty of training deep models with the binary constraint. \cite{8237860} proposed to handles the non-smooth problem of the $\sign$ function by continuation, i.e., starting the training with a smoothed approximation and gradually reducing the smoothness as the training proceeds, i.e., $\lim_{\beta\to\infty} \tanh(\beta x)=\sign(x)$. \cite{8578813} transformed the original binary optimization into differentiable optimization problem over hash functions through Taylor series expansion. \cite{Cao_2018_CVPR} introduced  a pairwise cross-entropy loss  based on the Cauchy distribution, which penalizes significantly similar image pairs
with Hamming distance larger than the given Hamming radius threshold, e.g., greater than 2.
Nevertheless, these methods requires class labels for the training process (i.e., supervised hashing). This is not the focus of our methods which aim to learn optimal binary codes from given image representations in the unsupervised manner.

}
\subsection{Contributions}
\red{
In this work, to address the problem of learning to preserve data affinity in low-dimension binary codes, \textit{(i)} we first propose a novel loss function to learn a single linear transformation under the \textit{column orthonormal constraint}\footnote{Please refer to section \ref{ssec:notations} for our term definitions.} in the unsupervised manner that {\em compresses} and {\em binarizes the input data jointly}. The approach is named as \textbf{Simultaneous Compression and Quantization (SCQ)}.
Noted that the idea of jointly compressing and binarizing data has been explored in \cite{BA,UH-BDNN}. However, due to the difficulty of the non-convex orthogonal constraint, these works try to relax the orthogonal constraint and resort to the reconstruction criterion as an indirect way to handle the similarity perserving concern.
Our work is the first one to tackle the similarity concern by enforcing \textbf{\textit{strict}} orthogonal constraints.  

\textit{(ii)} Under the strict orthogonal constraints, we conduct analysis and experiments to show that our formulation is able to retain a high amount of the variance, i.e., preserve data similarity, and achieve small quantization loss, which are important requirements in hashing for image retrieval (\cite{ITQ,BA,UH-BDNN}). As a result, this leads to improved accuracy as demonstrated in our experiments.
}



\textit{(iii)} We then propose to relax the \textit{column \textbf{orthonormal}} constraint to \textit{column \textbf{orthogonal}} constraint on the transformation matrix. The relaxation not only helps to gain extra retrieval performances but also significantly improves the training time.

\textit{(iv)} Our proposed loss functions, with column orthonormal and orthogonal constraints, are confronted with two main challenges. The first is the binary constraint, which is the traditional and well-known difficulty of hashing problem (\cite{LSH,ITQ,SH}). The second challenge is the non-convex nature of the orthonormal/orthogonal constraint (\cite{feasible_ortho_constraints}). To tackle the binary constraint, we propose to apply an alternating optimization with an auxiliary variable. Additionally, we resolve the orthonormal/orthogonal constraint by using the cyclic coordinate descent approach 
to learn one column of the projection matrix at a time while fixing the others. The proposed algorithms are named as \textit{Orthonormal Encoder (OnE)} and \textit{Orthogonal Encoder (OgE)}.

\textit{(v)} Comprehensive experiments on common benchmark datasets show considerable improvements on retrieval performance of proposed methods over other state-of-the-art hashing methods. 
Additionally, the computational complexity and training / online-processing time are also discussed to show the computational efficiency of our methods.


\subsection{Notations and Term definitions}
\label{ssec:notations}
We first introduce the notations. Given a zero-centered dataset $\mathbf{X}=\{\mathbf{x}_i\}_{i =1}^n\in \mathbb{R}^{n\times d}$ which consists of $n$ images and each image is represented by a $d$-dimension feature descriptor, our proposed hashing methods aim to learn {a column orthonormal/orthogonal matrix} $\mathbf{V} \in \mathbb{R}^{d\times L} (L \ll d)$ which simultaneously compresses input data $\mathbf{X}$ to $L$-dimensional space, while retains a high amount of variance, and quantizes to binary codes $\mathbf{B}\in \{-1,+1\}^{n\times L}$.

It is important to note that, in this work, we abuse the terms: \textbf{\textit{column orthonormal/orthogonal matrix}}. Specifically, the term \textbf{\textit{column orthonormal matrix}} is used to indicate the matrix $\mathbf{V}$ that $\mathbf{V}^\top\mathbf{V}=\mathbf{I}_{L\times L}$, where $\mathbf{I}_{L\times L}$ is the $L\times L$ identity matrix. While the term \textbf{\textit{column orthogonal matrix}} indicates matrix $\mathbf{V}$ that $\mathbf{V}^\top\mathbf{V}=\mathbf{D}_{L\times L}$, where $\mathbf{D}_{L\times L}$ is an arbitrary $L\times L$ diagonal matrix. 
Noted that the word ``{column}'' in these terms means that columns of the matrix are pairwise independent.

We define $\Lambda=[\lambda_1,\cdots, \lambda_d]$ as the eigenvalues of the covariance matrix $\mathbf{X}^T\mathbf{X}$ sorted in descending order. 
Finally, let $\mathbf{b}_k, \mathbf{v}_k$ be the $k$-th $(1\le k\le L)$ columns of $\mathbf{B}, \mathbf{V}$ respectively.

\smallskip
The remainder of the paper is organized as follow. Firstly, Section \ref{sec:orthonormal} presents in details our proposed hashing method, i.e., \textbf{Orthonormal Encoder (OnE)} and provide the analysis to show that our method can retain a high amount of variance and achieve small quantization loss. Section \ref{sec:OgE} presents a relax version of OnE, i.e., \textbf{Orthogonal Encoder (OgE)}. Section \ref{sec:exp} presents experiment results to validate the effectiveness of our proposed methods. We conclude the paper in Section \ref{sec:conclusion}.
\section{Simultaneous Compression \& Quantization: Orthonormal Encoder}\label{sec:orthonormal}

\subsection{Problem Formulation}
In order to jointly learn data dimension reduction and binary quantization using a single linear transformation $\mathbf{V}$, we propose to solve the following constrained optimization:
\begin{equation}\label{eq:OnE}
\begin{split}
&\arg\min\limits_{\mathbf{B},\mathbf{V}}\mathcal{Q}(\mathbf{B}, \mathbf{V}) = \frac{1}{n}\|\mathbf{B}-\mathbf{X}\mathbf{V}\|_F^2 \\ 
&\textrm{s.t. } \mathbf{V}^\top\mathbf{V} = \mathbf{I}_{L\times L}; \mathbf{B}\in \{-1, +1\}^{n\times L},
\end{split}
\end{equation}
where $\|\cdot\|_F$ denotes the Frobenius norm.
\red{
Additionally, the orthonormal constrained on the column of $\V$ is necessary to make sure no redundant information is captured in binary codes (\cite{SemiSupHash}) (i.e., the projected low-dimensional features are strictly pairwise uncorrelated) and the projection vectors do not scale up/down projected data. 
}

\red{
It is noteworthy to highlight the differences between our loss function Eq. (\ref{eq:OnE}) and the binary quantization loss function of ITQ (\cite{ITQ}).
Firstly, different from ITQ, which works on the compressed low-dimensional feature space after using PCA, i.e., $\X \in \mathbf{R}^{n\times L}$; our approach, instead, works directly on the original high-dimensional feature space $\X \in \mathbf{R}^{n\times d} (d\gg L)$. 
This leads to the second main difference that the non-square column orthonormal matrix $\mathbf{V} \in \mathbb{R}^{d\times L}$ simultaneously \textit{(i)} \textit{compresses data} to low-dimension and \textit{(ii)} \textit{quantizes to binary codes}.
However, it is important to note that solving for a non-square projection matrix $\V$ is challenging. To handle this difficulty,
ITQ propose to solve the data compression and binary quantization problems in two separated optimizations. Specifically, it applys PCA to compress data to $L$ dimension, and then uses the Orthogonal Procrustes approach (\cite{orthogonalProcrustes}) to learn a $L\times L$ square rotation matrix to optimize binary quantization loss.
However, there is a limitation in ITQ approach as no consideration for the binary constraint in the data compression step, i.e., PCA. Consequently, the solution is suboptimal. In this paper, by adopting recent advance in cyclic coordinate descent approach (\cite{SDH,UH-BDNN,CCDvsRCD,EPM-MPEC}), we propose a novel and efficient algorithm to resolve the ITQ limitation by simultaneously attacking both problems in a single optimization problem under the strict orthogonal constraint. Hence, our optimization can lead to a better optimal solution.

}

\subsection{Optimization}
\label{ssec:OnE_opt}
In this section, we discuss the key details of the algorithm (Algorithm \ref{algo1}) for solving the optimization problem Eq. (\ref{eq:OnE}). 
In order to handle the binary constraint in Eq. (\ref{eq:OnE}), we propose to use alternating optimization over $\mathbf{V}$ and $\mathbf{B}$.
\begin{algorithm}[t]
\caption{Orthonormal Encoder}\label{algo1}
\begin{flushleft}
\hspace*{\algorithmicindent}\textbf{Input:} \\
\hspace*{\algorithmicindent}\hspace*{\algorithmicindent} $\mathbf{X} = \{\mathbf{x}\}_{i=1}^n\in \mathbb{R}^{n\times d}$: training data; \\
\hspace*{\algorithmicindent}\hspace*{\algorithmicindent} $L$: code length; \\
\hspace*{\algorithmicindent}\hspace*{\algorithmicindent} $max\_iter$: maximum iteration number; \\
\hspace*{\algorithmicindent}\hspace*{\algorithmicindent} $\{\epsilon, \epsilon_b, \epsilon_u\}$: convergence error-tolerances; \\
\hspace*{\algorithmicindent}\textbf{Output} \\
\hspace*{\algorithmicindent}\hspace*{\algorithmicindent} Column Orthonormal matrix $\mathbf{V}$.
\end{flushleft}
\begin{algorithmic}[1]

\State Randomly initialize $\V$ such that $\V^\top\V=\I$.
\For {$t=1 \to max\_iter$} 
\Procedure{  Fix $\mathbf{V}$, update $\mathbf{B}$.}{}
\State Compute $\mathbf{B}$ (Eq. (\ref{eq:computeB})).
\EndProcedure
\Procedure{  Fix $\mathbf{B}$, update $\mathbf{V}$.}{}
\State Find $\nu_1$ using binary search (BS) (Eq. (\ref{eq:nu_cond})). 
\State Compute $\mathbf{v}_1$ (Eq. (\ref{eq:OnE_v1})).
\For {$k=2\to L$}
\Procedure {Solve $\mathbf{v}_k$}{}
\State Initialize $\Phi_k=[0, \cdots, 0]$.
\While {true}
\State Fix $\Phi_k$, solve for $\nu_k$ using BS.
\State Fix $\nu_k$, compute $\Phi_k=A_k^{-1}c_k$.
\State Compute $\mathbf{v}_k$ (Eq. (\ref{eq:vk})).
\If {$(\mathbf{v}_k^\top\mathbf{v}_k-1)< \epsilon_u$} 
\State \Return  $\mathbf{v}_k$
\EndIf
\EndWhile
\EndProcedure

\EndFor
\EndProcedure	

\If {$t>1$ and ${(\mathcal{Q}_{t-1}-\mathcal{Q}_{t})}/{\mathcal{Q}_{t}}< \epsilon$} 
\textbf{break}
\EndIf
\EndFor
\State \Return $\mathbf{V}$

\end{algorithmic}
\end{algorithm}

\subsubsection{Fix $\V$ and update $\mathbf{B}$} When $\mathbf{V}$ is fixed, the problem becomes exactly the same as when fixing rotation matrix in ITQ. To make the paper self-contained, we repeat the explaination of \cite{ITQ}. By expanding the objective function in Eq. (\ref{eq:OnE}), we have
\begin{equation}
\label{eq:fixV}
\begin{split}
\mathcal{Q}(\B, \V) = &\frac{1}{n}\left(\|\B\|_F^2 + \|\U\|_F^2-2\textrm{tr}(\B\U)\right) \\
& =\frac{1}{n}\left(nL + \|\U\|_F^2-2\textrm{tr}(\B\U)\right),
\end{split}
\end{equation}
where $\U=\X\V$. 
Because $\V$ is fixed, so $\U$ is fixed, minimizing (\ref{eq:fixV}) is equivalent to maximizing
\begin{equation}
\textrm{tr}(\B\U) = \sum_{i=1}^n\sum_{j=1}^LB_{ij}U_{ij}
\end{equation}
where $B_{ij}$ and $U_{ij}$ denotes elements of $\B$ and $\U$ respectively.
To maximize this expression with respect to $\B$, we need to have $B_{ij} = 1$ whenever $U_{ij}\ge 0$ and $B_{ij} = -1$ otherwise. 
Hence, the optimal value of $\mathbf{B}$ can be simply achieved by
\begin{equation}\label{eq:computeB}
\mathbf{B}={sign}(\X\V).
\end{equation}

\subsubsection{Fix $\mathbf{B}$ and update $\V$} When fixing $\mathbf{B}$, the optimization is no longer a mix-integer problem. However, the problem is still non-convex and difficult to solve due to the orthonormal constraint (\cite{feasible_ortho_constraints}). It is important to note that $\mathbf{V}$ is not a square matrix. It means that the objective function is not the classic Orthogonal Procrustes problem (\cite{orthogonalProcrustes}). Hence, we cannot achieve the closed-form solution for $\mathbf{V}$ as proposed in \cite{ITQ}. To the best of our knowledge, there is no easy way for achieving the closed-form solution of non-square $\mathbf{V}$. Hence, in order to overcome this challenge, inspired by PCA and recent methods in cyclic coordinate descent (\cite{SDH,UH-BDNN,CCDvsRCD,EPM-MPEC}), we iteratively learn one vector, i.e., one column of $\mathbf{V}$, at a time. 
We now consider two cases for $k=1$ and $2\le k \le L$. 
\begin{itemize}
\item \textbf{$1$-st vector}
\end{itemize}
\begin{equation}
\arg\min\limits_{\mathbf{v}_1}\mathcal{Q}_1 = \frac{1}{n}\|\mathbf{b}_1-\mathbf{X}\mathbf{v}_1\|^2\quad\textrm{s.t. }\mathbf{v}_1^\top\mathbf{v}_1=1,
\end{equation}
where $\|\cdot\|$ is the $l2$-norm.

Let $\nu_1\in \mathbb{R}$ be the Lagrange multiplier, we formulate the \textit{Lagrangian} $\mathcal{L}_1$:
\begin{equation}\label{eq:Lagrange1}
\mathcal{L}_1(\mathbf{v}_1, \nu_1) = \frac{1}{n}\|\mathbf{b}_1-\mathbf{X}\mathbf{v}_1\|^2+\nu_1(\mathbf{v}_1^\top\mathbf{v}_1-1).
\end{equation}
By minimizing $\mathcal{L}_1$ over $\mathbf{v}_1$, we can achieve:
\begin{equation} \label{eq:OnE_v1}
\mathbf{v}_1 =(\mathbf{X}^\top\mathbf{X}+n\nu_1 \mathbf{I})^{-1}\mathbf{X}^\top\mathbf{b}_1,
\end{equation}
given $\nu_1$ that maximizes the \textit{dual function} $\mathcal{G}_1(\nu_1)$~\footnote{The dual function $\mathcal{G}_1(\nu_1)$ can be simply constructed by substituting $\mathbf{v}_1$ from Eq. (\ref{eq:OnE_v1}) into Eq. (\ref{eq:Lagrange1}).} of $\mathcal{L}_1(\mathbf{v}_1, \nu_1)$ (\cite{cvxbook}). Equivalently, $\nu_1$ should satisfy the following conditions:
\begin{equation}\label{eq:nu_cond}
\begin{cases}
\nu_1 > -\lambda_d/n \\
\begin{split}
\frac{\partial\mathcal{G}_1}{\partial\nu_1} = [(\mathbf{X}^\top\mathbf{X}+n\nu_1 \mathbf{I})^{-1}\mathbf{X}^\top\mathbf{b}_1]^\top& \\
[(\mathbf{X}^\top\mathbf{X}+n\nu_1 \mathbf{I})^{-1}\mathbf{X}^\top\mathbf{b}_1&] - 1 = 0
\end{split}
\end{cases}
\end{equation}
where $\lambda_d$ is the smallest eigenvalue of $\mathbf{X}^\top\mathbf{X}$. The detail derivation is provided in Appendix section \ref{sec:derivation_1}.

In Eq. (\ref{eq:nu_cond}), the first condition is to ensure that $(\mathbf{X}^\top\mathbf{X}+n\nu_1 \mathbf{I})$ is non-singular and the second condition is achieved by setting the derivative of $\mathcal{G}_1(\nu_1)$ with regard to $\nu_1$ equal to $0$.

The second equation in Eq. (\ref{eq:nu_cond}) can be recognized as a $d$-order polynomial equation of $\nu_1$ which has no explicit closed-form solution for $\nu_1$ when $d > 4$. Fortunately, since $\mathcal{G}(\nu_1)$ is a concave function of $\nu_1$, ${\partial\mathcal{G}_1}/{\partial\nu_1}$ is monotonically decreasing. Hence, 
we can simply solve for $\nu_1$ using binary search with a small error-tolerance $\epsilon_b$. Note that:
\begin{equation}
\label{eq:lim_dual_func}
\begin{cases}
\lim\limits_{\nu_1\to (-\lambda_d/n)^+}\frac{\partial\mathcal{G}_1}{\partial\nu_1} = + \infty \\
\lim\limits_{\nu_1\to +\infty}\frac{\partial\mathcal{G}_1}{\partial\nu_1} = -1
\end{cases},
\end{equation}
thus ${\partial\mathcal{G}_1}/{\partial\nu_1}=0$ always has a solution.

\begin{itemize}
\item \textbf{$k$-th vector $(2\le k \le L)$}
\end{itemize}
For the second vector onward, besides the unit-norm constraint, we also need to ensure that the current vector is independent with its $(k-1)$ previous vectors.
\begin{equation}
\begin{split}
&\arg\min_{\mathbf{v}_k}\mathcal{Q}_k = \frac{1}{n}\|\mathbf{b}_k-\mathbf{X}\mathbf{v}_k\|^2\\
\textrm{s.t. }&\mathbf{v}_k^\top\mathbf{v}_k=1; \mathbf{v}_k^\top\mathbf{v}_i=0, \forall i\in [1,k-1].
\end{split}
\end{equation}

Let $\nu_k\in \mathbb{R}$ and $\Phi_k = [\phi_{k1}, ..., \phi_{k(k-1)}]^\top\in \mathbb{R}^{(k-1)}$ be the Lagrange multipliers, we also formulate the \textit{Lagrangian} $\mathcal{L}_k$:
\begin{equation}
\label{eq:OnE_vk}
\begin{split}
\mathcal{L}_k(\mathbf{v}_k, \nu_k, \Phi_k) &= \frac{1}{n}\|\mathbf{b}_k-\mathbf{X}\mathbf{v}_k\|^2 \\
&+\nu_k(\mathbf{v}_k^\top\mathbf{v}_k-1) +\sum_{i=1}^{k-1}\phi_{ki}\mathbf{v}_k^\top\mathbf{v}_i.
\end{split}
\end{equation}
Minimizing $\mathcal{L}_k$ over $\mathbf{v}_k$, similar to Eq. (\ref{eq:OnE_v1}), we can achieve:
\begin{equation}\label{eq:vk}
\mathbf{v}_k=(\mathbf{X}^\top\mathbf{X}+n\nu_k \mathbf{I})^{-1}\left(\mathbf{X}^\top\mathbf{b}_k-\frac{n}{2}\sum_{i=1}^{k-1}\phi_{ki}\mathbf{v}_i\right),
\end{equation}
given $\{\nu_k, \Phi_k\}$ that satisfy the following conditions which make the corresponding \textit{dual function} $\mathcal{G}_k(\nu_k, \Phi_k)$ maximum:
\begin{equation}
\label{eq:nu_k_cond}
\begin{cases}
\nu_k > -\lambda_d/n \\
\mathbf{v}_k^\top\mathbf{v}_k = 1\\
\mathbf{A}_k\Phi_k = \mathbf{c}_k
\end{cases}
\end{equation}
where
\begin{equation}
\begin{cases}
\mathbf{A}_k = \frac{n}{2}\begin{bmatrix} \mathbf{v}_1^\top\mathbf{Z}_k\mathbf{v}_1 & \cdots & \mathbf{v}_1^\top\mathbf{Z}_k\mathbf{v}_{(k-1)} \\
\vdots & \ddots & \vdots \\
\mathbf{v}_{(k-1)}^\top\mathbf{Z}_k\mathbf{v}_1 & \cdots &  \mathbf{v}_{(k-1)}^\top\mathbf{Z}_k\mathbf{v}_{(k-1)}
\end{bmatrix}\\
\mathbf{c}_k = \begin{bmatrix}
\mathbf{v}_1^\top\mathbf{Z}_k\mathbf{X}^\top\mathbf{b}_k&
\cdots &
\mathbf{v}_{(k-1)}^\top\mathbf{Z}_k\mathbf{X}^\top\mathbf{b}_k
\end{bmatrix}^\top
\end{cases}
\end{equation}
in which $\mathbf{Z}_k = (\mathbf{X}^\top\mathbf{X}+n\nu_k \mathbf{I})^{-1}$. The detail derivation is provided in Appendix section \ref{sec:derivation_2}.

There is also no straight-forward solution for $\{\nu_k, \Phi_k\}$. In order to resolve this difficulty, we propose to use alternative optimization to solve for $\nu_k$ and $\Phi_k$. In particular, (i) given a fixed $\Phi_k$ (initialized as $[0, \cdots, 0]^\top$), we find $\nu_k$ using binary search as discussed above. Additionally, similar to $\nu_1$, there is always a solution for $\nu_k$. Then, (ii) with fixed $\nu_k$, we can get the closed-form solution for $\Phi_k$ as $\Phi_k=\mathbf{A}_k^{-1}\mathbf{c}_k$. 
Note that since the dual function $\mathcal{G}_k$ is a concave function of $\{\nu_k, \Phi_k\}$, alternative optimizing between $\nu_k$ and $\Phi_k$ still guarantees the solution to approach the global optimal one. 

\red{
Additionally, we note that solving for $\{\nu_k, \Phi_k\}$ requires a matrix inversion $\mathbf{Z}_k^{-1}$ (for each $\nu_k$), which is very computationally expensive. However, by utilizing the Singular Value Decomposition (SVD), we can efficiently compute the inversion as follows:
\begin{equation}
(\mathbf{X}^\top\mathbf{X} + n\nu_k \mathbf{I})^{-1} = \mathbf{\hat{U}}\mathbf{\hat{\Sigma}} \mathbf{\hat{U}}^\top,
\end{equation}
where $\mathbf{\hat{U}}\in\mathbb{R}^{d\times d}$ is the matrix of eigenvectors corresponding to 
$\hat{\Lambda}$ in columns ($\hat{\Lambda} = [\lambda_d, \cdots, \lambda_1]$ is the eigenvalues of $\mathbf{X}^\top\mathbf{X}$ sorted in \textit{ascending} order) and
\begin{equation}
    \mathbf{\hat{\Sigma}}=\textrm{diag}\left(\left[\frac{1}{\lambda_d+n\nu_k}, \cdots, \frac{1}{\lambda_1+n\nu_k}\right]\right)
\end{equation}
with ``$\textrm{diag}(\cdot)$''  is the operation to convert vectors to square diagonal matrices. Note that, given $\mathbf{X}$, $\mathbf{\hat{U}}$  and $\hat{\Lambda}$ are fixed and can be computed in advance. 
}

\begin{figure}[t]
\centering
\includegraphics[width=0.33\textwidth]{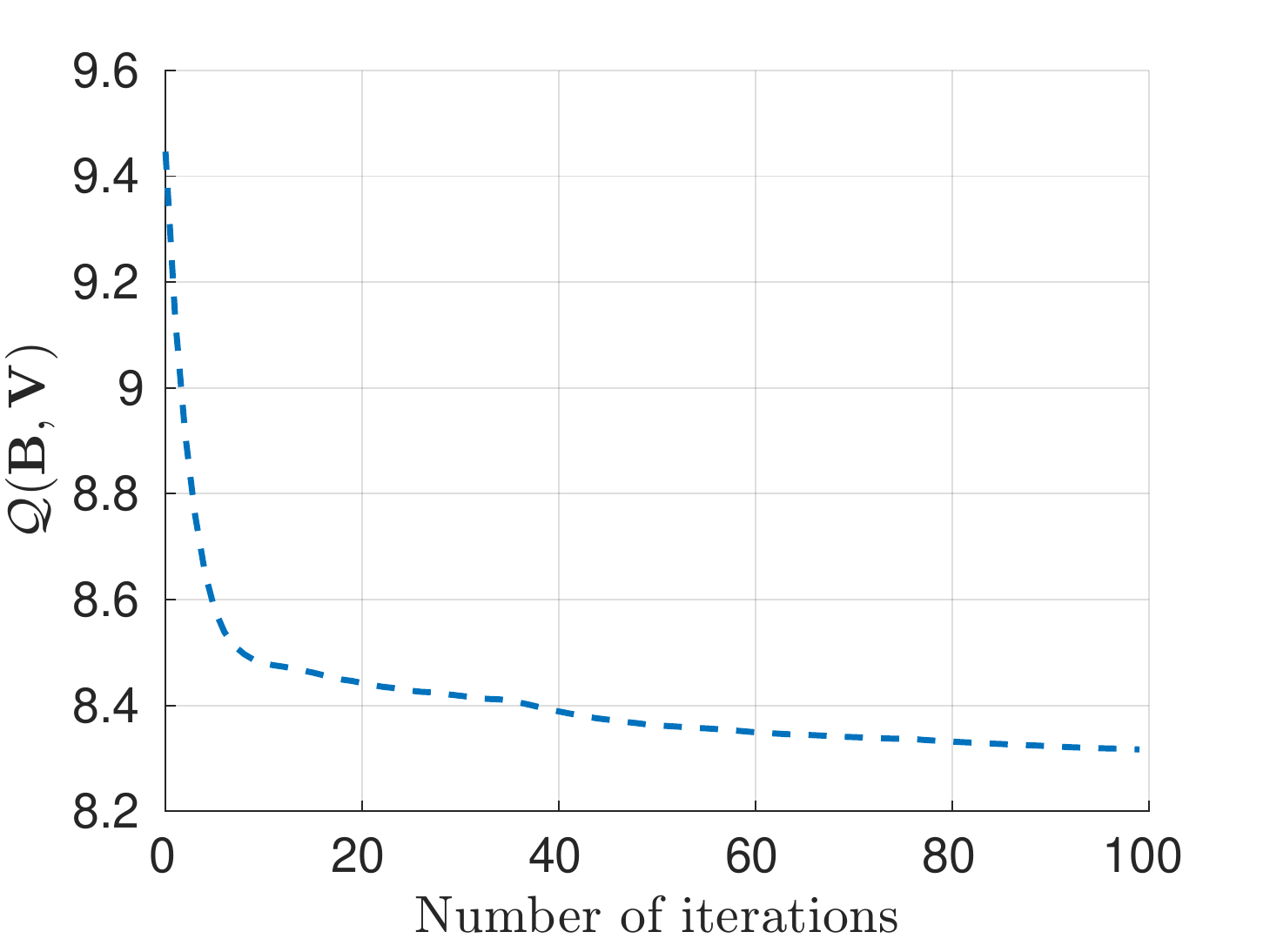}
\caption{Quantization error for learning the projection matrix $\V$ with $L=32$ on the CIFAR-10 dataset (section \ref{ssec:datasets}).}
\label{fig:loss_by_iter}
\end{figure}

\smallskip
Figure \ref{fig:loss_by_iter} shows an error convergence curve of the optimization problem Eq. (\ref{eq:OnE}).
 We stop the optimization when the relative reduction of the quantization loss is less than $\epsilon$, i.e., ${(\mathcal{Q}_{t-1}-\mathcal{Q}_{t})}/{\mathcal{Q}_{t}}< \epsilon$.

\subsection{Retained variance and quantization loss}
\label{ssec:var_vs_quan}

In the hashing problem for image retrieval, both retained variance and quantization loss are important.
In this section, we provide analysis to show that, when solving Eq. (\ref{eq:OnE}), it is possible to retain a high amount of the variance and achieve small quantization loss.
As will be discussed in more details, this can be accomplished by applying an appropriate scale S on the input dataset. 
Noticeably, by applying any positive scale $\Sc > 0$~\footnote{For simplicity, we only discuss positive value $\Sc > 0$. Negative value $\Sc < 0$ should have similar effects.} on the dataset, the local structure of data is strictly preserved, i.e., the ranking nearest neighbor set of every data point is always the same. Therefore, in the hashing problem for retrieval task, it is equivalent to work on a scaled version of the dataset, i.e., $\X_\Sc = \Sc\X$.
We can re-write the loss function of Eq. (\ref{eq:OnE}) as following:
\begin{equation}\label{eq:rewriteOnE}
\mathcal{Q}(\Sc, \V)=\|\1 -\Sc|\X\V|\|_F^2\quad \textrm{s.t. } \V^\top\V = \mathbf{I}_{L\times L}; \Sc > 0,
\end{equation} where $|\cdot|$ is the element-wise operation to find the absolute values and $\1$ is the all-1 $(n\times L)$ matrix.
In what follows, we discuss how $\Sc$ can affect the retained variance and quantization loss.

\subsubsection{Maximizing retained variance}
\label{ssec:scale4var}

\begin{figure}[t]
\centering
\includegraphics[width=0.25\textwidth]{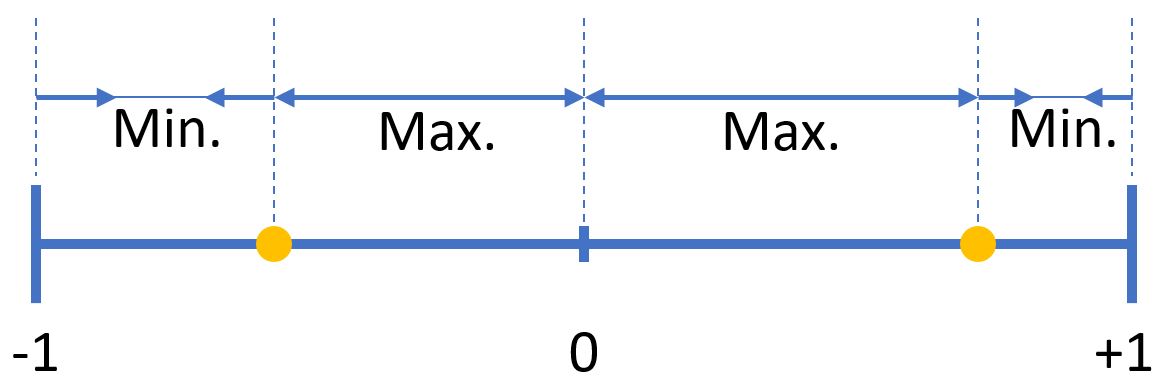}
\caption{An illustration of the relationship between the \textit{minimizing quantization loss} and \textit{maximizing retained variance} problems.}
\label{fig:illustration}
\end{figure}

We recognize that by scaling to the dataset $\mathbf{X}$ by an appropriate scale $\Sc$, such that all projected data points are inside the hyper-cube of $1$, i.e.,
$\max(\Sc|\X\V|) \le 1$, the \textit{maximizing retained variance problem} (PCA) can achieve similar results to the \textit{minimizing quantization loss problem}, i.e., 
$\arg\max\limits_\V\|\Sc\X\V\|_F^2 \approx \arg\min\limits_\V\|\1-\Sc|\X\V|\|_F^2$.
Intuitively, we can interpret the former problem, i.e., PCA, as to find the projection that maximizes the distances of projected data points from the coordinate origin. While the latter problem, i.e., minimizing binary quantization loss, tries to find the projection matrix that minimizes the distances of projected data points from $-1$ or $+1$ correspondingly. A simple 1-D illustration to explain the relationship between two problems is given in Figure \ref{fig:illustration}.

Since each vector of $\V$ is constrained to have the unit norm, the condition $\max(\Sc|\X\V|) \le 1$ actually can be satisfied by scaling the dataset by $\Sc_\text{max\_var}$ to have all data points in the original space inside the hyper-ball with unit radius, in which $1/\Sc_\text{max\_var}$ is equal to the largest $l2$-distance between data points and the coordinate origin.

\subsubsection{Minimizing quantization loss}
\label{ssec:scale4quantization}

\begin{figure*}[ht]
\centering
\begin{subfigure}[b]{0.32\textwidth}
\includegraphics[width=\textwidth]{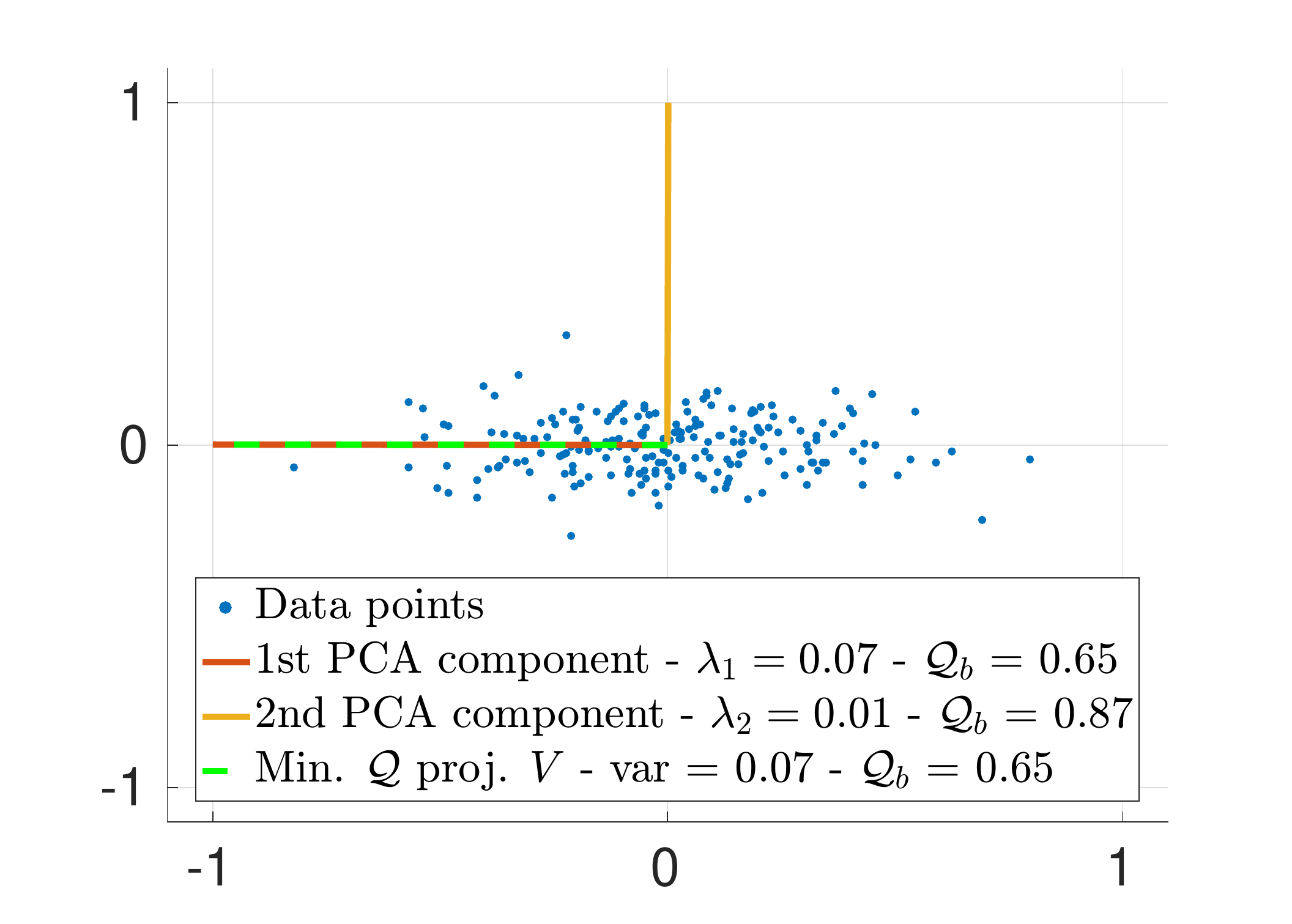}
\caption{$\Sc=\Sc_\textrm{max\_var}$}
\label{fig:toy_a}
\end{subfigure}
\begin{subfigure}[b]{0.32\textwidth}
\includegraphics[width=\textwidth]{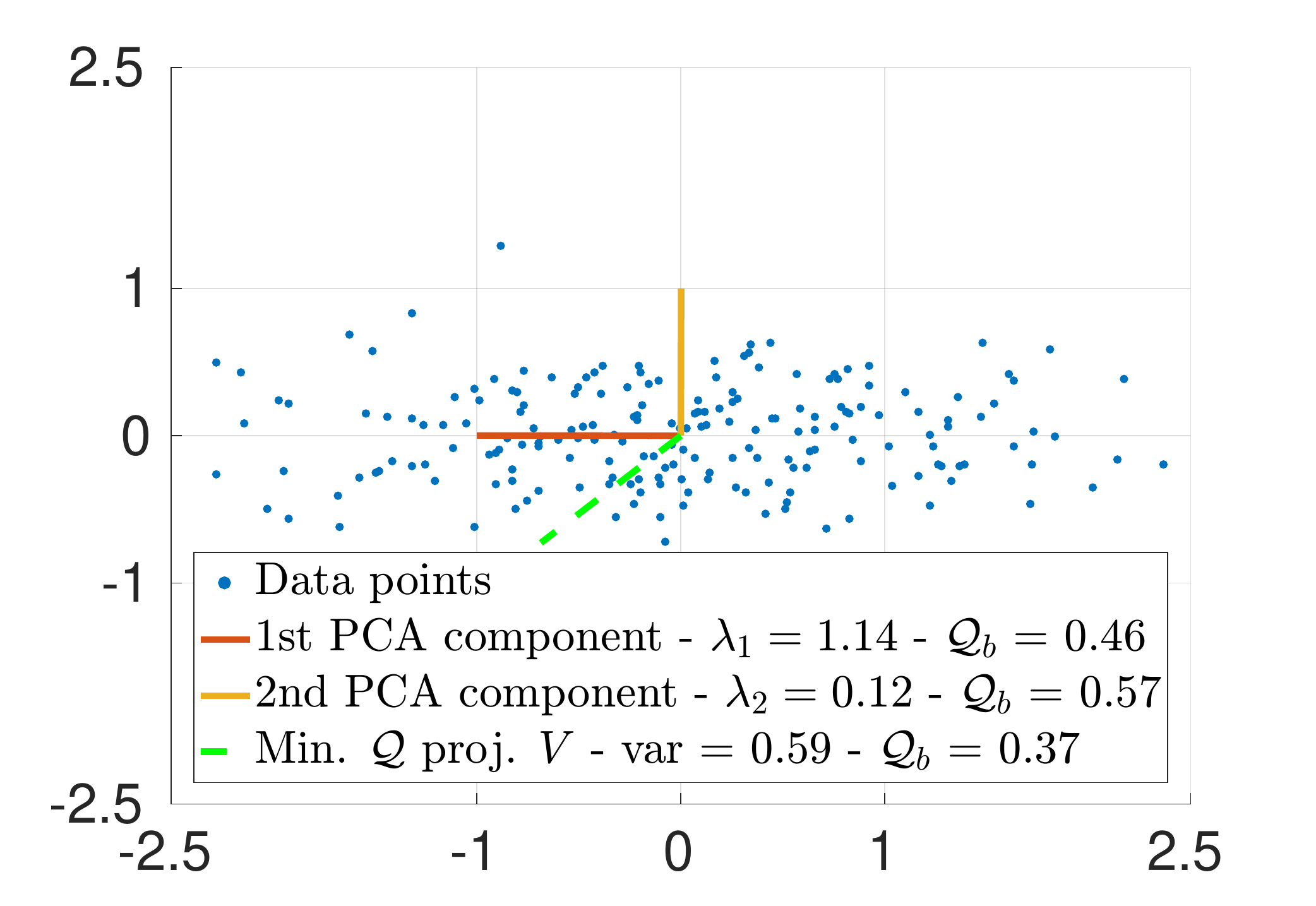}
\caption{$\Sc=4\Sc_\textrm{max\_var}$}
\label{fig:toy_c}
\end{subfigure}
\begin{subfigure}[b]{0.32\textwidth}
\includegraphics[width=\textwidth]{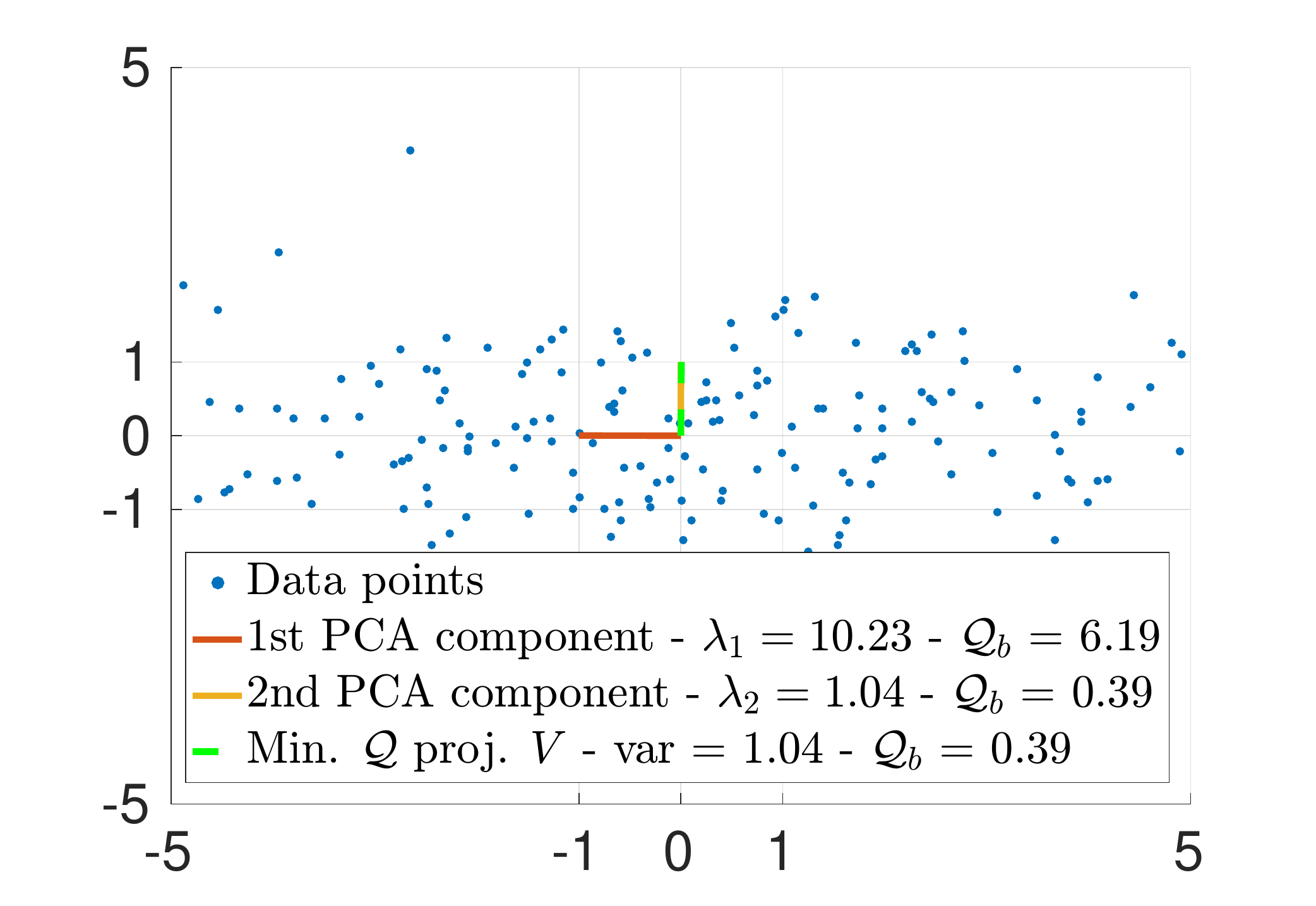}
\caption{$\Sc=12\Sc_\textrm{max\_var}$}
\label{fig:toy_d}
\end{subfigure}
\caption{A toy example for $n=200,d=2$ and $L=1$ to illustrate how the quantization loss and the minimizing quantization loss vector (green dash line) vary when $\Sc$ increases. The values in legends present the variances and the quantization losses per bit, $\mathcal{Q}_b$, of the data which is projected in corresponding vectors (rounding to two decimal places). 
}
\label{fig:toy_example}
\end{figure*}

Regarding the quantization loss $\mathcal{Q}(\Sc, \V)$ (Eq. \ref{eq:rewriteOnE}), which is a convex function of $\Sc|\X\V|$, 
by setting ${\partial \mathcal{Q}(\Sc, \V)}/{\partial \Sc|\X\V|}=\0$, we have the optimal solution for $\mathcal{Q}(\Sc, \V)$ as following:
\begin{equation} \label{eq:mean-var}
\frac{\partial \mathcal{Q}(\Sc, \V)}{\partial \Sc|\X\V|}=2(\Sc|\X\V|-\1)=\0 \Leftrightarrow\begin{cases}
\textrm{mean}(|\X\V|)=1/\Sc\\
\textrm{var}(|\X\V|)=0
\end{cases}
\end{equation}
where $\0$ is the all-0 $(n\times L)$ matrix.

Considering $\Sc\ge\Sc_\text{max\_var}$, there are two important findings.
Firstly, there is obviously no scaling value $\Sc$ that can concurrently achieve $\Sc\max(|\X\V|)\le 1$ and $\Sc~\textrm{mean}(|\X\V|)=1$, except the case $\Sc|\X\V|=\1$ which is unreal in practice.
Secondly, from Eq. (\ref{eq:mean-var}), we can recognize that as $\Sc$ gets larger,
i.e., $1/\Sc$ gets smaller, minimizing the loss $\mathcal{Q}$ will produce $\V$ that focuses on lower-variance directions so as to achieve smaller $\textrm{mean}(|\X\V|)$ as well as smaller $\textrm{var}(|\X\V|)$.
It means that $\Sc|\X\V|$ gets closer to the global minimum of $\mathcal{Q}(\Sc, \V)$. Consequently, the quantization loss becomes smaller. 
In Figure \ref{fig:toy_example}, we show a toy example to illustrate that as $\Sc$ increases, minimizing quantization loss diverts the projection vector from top-PCA component (Figure \ref{fig:toy_a}) to smaller variance directions (Figure \ref{fig:toy_c} $\to$ \ref{fig:toy_d}), while the quantization loss (per bit) gets smaller (Figure \ref{fig:toy_a} $\to$ \ref{fig:toy_d}). In summary, as $\textrm{var}(|\X\V|)$ gets smaller, the quantization loss is smaller and vice versa.
However, note that keeping increasing $\Sc$ when $\V$ already focuses on least-variance directions will make the quantization loss larger. 

\smallskip
Note that the scale $\Sc$ is a hyper-parameter in our system.
In the experiment section (Section \ref{ssec:effect_of_param}), we will additionally conduct experiments to quantitatively analyse the effect of the scale hyper-parameter $\Sc$ and 
determine proper values using validation dataset. 

%

\section{Simultaneous Compression \& Quantization: Orthogonal Encoder}
\label{sec:OgE}

\subsection{Problem Re-formulation: Orthonormal to Orthogonal}
\label{ssec:OgE_problem_form}

In Orthonormal Encoder (OnE), we work with the \textit{column orthonormal} constraint on  $\mathbf{V}$. However, we recognize that relaxing this constraint to \textit{column orthogonal} constraint, i.e., relaxing the unit norm constraint on each column of $\V$, by converting it into a penalty term, provides three important advantages. We now achieve the new loss function as following:
\begin{equation}\label{eq:OgE}
\begin{split}
\arg &\min\limits_{\mathbf{B}, \mathbf{V}}\mathcal{Q}(\mathbf{B}, \mathbf{V}) = \frac{1}{n}\|\mathbf{B}-\mathbf{XV}\|_F^2 + \mu\sum_{i=1}^L\mathbf{v}_i^\top\mathbf{v}_i \\ 
&\textrm{s.t. } \mathbf{B}\in \{-1, +1\}^{n\times L}; \mathbf{v}_i^\top\mathbf{v}_j = 0, \forall i\neq j,
\end{split}
\end{equation}
where $\mu$ is a fixed positive hyper-parameter to penalize large norms of $\mathbf{v}_i$. It is important to note that, in Eq. (\ref{eq:OgE}), we still enforce the strict pairwise independent constraint of projection vectors to  ensure no redundant information is captured.

Firstly, with an appropriately large $\mu$, the optimization prefers to choose large variance components of $\X$ since this helps to achieve the projection vectors that have smaller norms. In other words, without penalizing large norms of $\mathbf{v}_i$, the optimization has no incentive to focus on high variance components of $\X$ since it can produce projection vectors with arbitrary large norms that can scale any components appropriately to achieve minimum binary quantization loss.
Secondly, this provides more flexibility of having different scale values for different directions.
Consequently, relaxing the unit-norm constraint of each column of $\V$ helps to mitigate the difficulty of choosing the scale value $\Sc$. 
However, it is important to note that a too large $\mu$, on the other hand, may distract the optimization from minimizing the binary quantization term.  
Finally, from OnE Optimization (Section \ref{ssec:OnE_opt}), we observed that the unit norm constraint on each column of $\mathbf{V}$ makes the OnE optimization difficult to be solved efficiently since there is no closed-form solution for $\{\mathbf{v}\}_{k=1}^L$. By relaxing this unit norm constraint, we now can achieve the closed-form solutions for $\{\mathbf{v}\}_{k=1}^L$; hence, it is very computationally beneficial. We will discuss more about the computational aspect in section \ref{sssec:complexity}. 


\subsection{Optimization}
\label{ssec:OgE_opt}

Similar to the Algorithm \ref{algo1} for solving {Orthonormal Encoder}, we apply alternative optimize $\mathbf{V}$ and $\mathbf{B}$ with the $\mathbf{B}$ step is exactly the same as Eq. (\ref{eq:computeB}). For $\mathbf{V}$ step, we also utilize the cyclic coordinate descent approach to iteratively solve $\mathbf{V}$, i.e., column by column. The loss functions are rewritten and their corresponding closed-form solutions for $\{\mathbf{v}\}_{k=1}^L$ can be efficiently achieved as following:
\begin{itemize}
\item \textbf{$1$-st vector}
\end{itemize}
\begin{equation}
\label{eq:OnE_1}
\arg\min\limits_{\mathbf{v}_1}\mathcal{Q}_1 = \frac{1}{n}\|\mathbf{b}_1-\mathbf{X}\mathbf{v}_1\|^2+\mu\mathbf{v}_1^\top\mathbf{v}_1.
\end{equation}
We can see that Eq. (\ref{eq:OnE_1}) is the regularized least squares problem, whose closed-form solution is given as:
\begin{equation}\label{eq:v11}
\mathbf{v}_1 = (\mathbf{X}^\top\mathbf{X}+n\mu \mathbf{I})^{-1}\mathbf{X}^\top\mathbf{b}_1.
\end{equation}

\begin{itemize}
\item \textbf{$k$-th vector $(2\le k\le L)$}
\end{itemize}
\begin{equation}
\begin{split}
\arg &\min_{\mathbf{v}_k}\mathcal{Q}_k = \frac{1}{n}\|\mathbf{b}_k-\mathbf{X}\mathbf{v}_k\|^2+\mu\mathbf{v}_k^\top\mathbf{v}_k\\
&\textrm{s.t. }\mathbf{v}_k^\top\mathbf{v}_i=0, \forall i\in [1,k-1].
\end{split}
\end{equation}
Given the Lagrange multiplier $\Phi_k = [\phi_{k1},\cdots,\phi_{k(k-1)}]^\top\in \mathbb{R}^{(k-1)}$, similar to Eq. (\ref{eq:OnE_v1}) and Eq. (\ref{eq:OnE_vk}), we can obtain $\mathbf{v}_k$ as following:
\begin{equation}\label{eq:vk1}
\mathbf{v}_k=(\mathbf{X}^\top\mathbf{X}+n\mu \mathbf{I})^{-1}\left(\mathbf{X}^\top\mathbf{b}_k-\frac{n}{2}\sum_{i=1}^{k-1}\phi_{ki}\mathbf{v}_i\right),
\end{equation}
where $\Phi_k = \mathbf{A}_k^{-1}\mathbf{c}_k$, in which
\begin{equation}
\begin{cases}
\mathbf{A}_k = \frac{n}{2}\begin{bmatrix} \mathbf{v}_1^\top\mathbf{Z}\mathbf{v}_1 & \cdots & \mathbf{v}_1^\top\mathbf{Z}\mathbf{v}_{(k-1)} \\
\vdots & \ddots & \vdots \\
\mathbf{v}_{(k-1)}^\top\mathbf{Z}\mathbf{v}_1 & \cdot &  \mathbf{v}_{(k-1)}^\top\mathbf{Z}\mathbf{v}_{(k-1)}
\end{bmatrix}\\
\mathbf{c}_k = \begin{bmatrix}
\mathbf{v}_1^\top\mathbf{Z}\mathbf{X}^\top\mathbf{b}_k &
\cdots &
\mathbf{v}_{(k-1)}^\top\mathbf{ZX}^\top\mathbf{b}_k
\end{bmatrix}^\top
\end{cases}
\end{equation}
and $\mathbf{Z} = (\mathbf{X}^\top\mathbf{X}+n\mu \mathbf{I})^{-1}$. 

Note that, given a fixed $\mu$, $\mathbf{Z}$ is a constant matrix, the $(k-1)\times (k-1)$ matrix $\mathbf{A}_{k}$ contains $(k-2)\times (k-2)$ matrix $\mathbf{A}_{(k-1)}$ in the top-left corner. It means that only the $(k-1)$-th row and column of matrix $\mathbf{A}_k$ are needed to be computed. Thus, $\Phi_k$ can be solved even more effectively. 

Finally, similar to OnE (Fig. \ref{fig:loss_by_iter}), we also empirically observe the convergence of the optimization problem Eq. \ref{eq:OgE}. We summarize the Orthogonal Encoder method in Algorithm \ref{algo2}. 

\begin{algorithm}[t]
\caption{Orthogonal Encoder}\label{algo2}
\begin{flushleft}
\hspace*{\algorithmicindent} \textbf{Input:} \\
\hspace*{\algorithmicindent} \hspace*{\algorithmicindent} $\mathbf{X} = \{\mathbf{x}\}_{i=1}^n\in \mathbb{R}^{n\times d}$: training data; \\
\hspace*{\algorithmicindent} \hspace*{\algorithmicindent} $L$: code length; \\
\hspace*{\algorithmicindent} \hspace*{\algorithmicindent}$max\_iter$: maximum iteration number; \\
\hspace*{\algorithmicindent} \hspace*{\algorithmicindent} $\epsilon$: convergence error-tolerance; \\
\hspace*{\algorithmicindent} \textbf{Output} \\
\hspace*{\algorithmicindent} \hspace*{\algorithmicindent} Column Orthogonal matrix $\mathbf{V}$.
\end{flushleft}
\begin{algorithmic}[1]

\State Randomly initialize $\V$ such that $\V^\top\V=\I$.
\For {$t=1 \to max\_iter$} 
\State Fix $\V$, update $\B$: Compute $\B$ (Eq. (\ref{eq:computeB})).
\State Fix $\B$, update $\V$: Compute $\{\mathbf{v}_i\}_{i=1}^L$ (Eq. (\ref{eq:v11}), (\ref{eq:vk1})).	
\If {$t>1$ and ${(\mathcal{Q}_{t-1}-\mathcal{Q}_{t})}/{\mathcal{Q}_{t}}< \epsilon$} 
\textbf{break}
\EndIf
\EndFor
\State \Return $\V$

\end{algorithmic}
\end{algorithm}

\subsection{Complexity analysis}
\label{sssec:complexity}
The complexity of the two algorithms, OnE and OgE, are shown in Table \ref{tb:complexity}. In our empirical experiments, $t$ is usually around 50, $t_1$ is at most 10 iterations, and $d^2\gg n$ (for CNN fully-connected features (Section \ref{ssec:datasets})). Firstly, we can observe that OgE is very efficient as its complexity is only linearly depended on the number of training samples $n$, feature dimension $d$, and code length $L$. In addition, OgE is also faster than OnE.
\red{Furthermore, as our methods aim to learn the projection matrices that preserve high-variance components, it is unnecessary to work on very high dimensional features. As there are many low-variance/noisy components, which will be discarded eventually. More importantly, we observe no retrieval performance drop when applying PCA to compress features to a much lower dimension, e.g., 512-D, in comparison with using the original 4096-D features. While this helps to achieve significant speed-up in training time for both algorithms, especially for the OnE, as its time complexity 
is depended on $d^3$ for large $d$.}
In addition, we conduct experiments to measure the actual running time of the algorithms and compare with other methods in section \ref{ssec:train_time}.
\begin{table}[t]
\caption{Computational complexity of algorithm OnE and OgE.
where $n$ is the number of training samples, $d$ is the feature dimension, $t$ is the number of iteration to alternative update $\B$ and $\V$, and $t_1$ is the number of iterations for solving $\mathbf{v}_k$ in Algorithm \ref{algo1}.
}
\label{tb:complexity}
\centering
\begin{tabular}{|c|c|}
\hline
&  Computational complexity\\
\hline
OnE & $\mathcal{O}(tt_1dL(\max (n,d^2) ))$\\ 
\hline
OgE & $\mathcal{O}(tdLn)$\\ 
\hline
\end{tabular}
\end{table}


\section{Experiments}
\label{sec:exp}
\subsection{Datasets, Evaluation protocols, and Implementation notes} 
\label{ssec:datasets}

\begin{figure*}[t]
\centering
\begin{subfigure}[b]{0.8\textwidth}
\includegraphics[width=\textwidth]{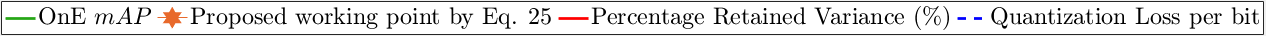}
\end{subfigure}

\begin{subfigure}[b]{0.32\textwidth}
\includegraphics[width=\textwidth]{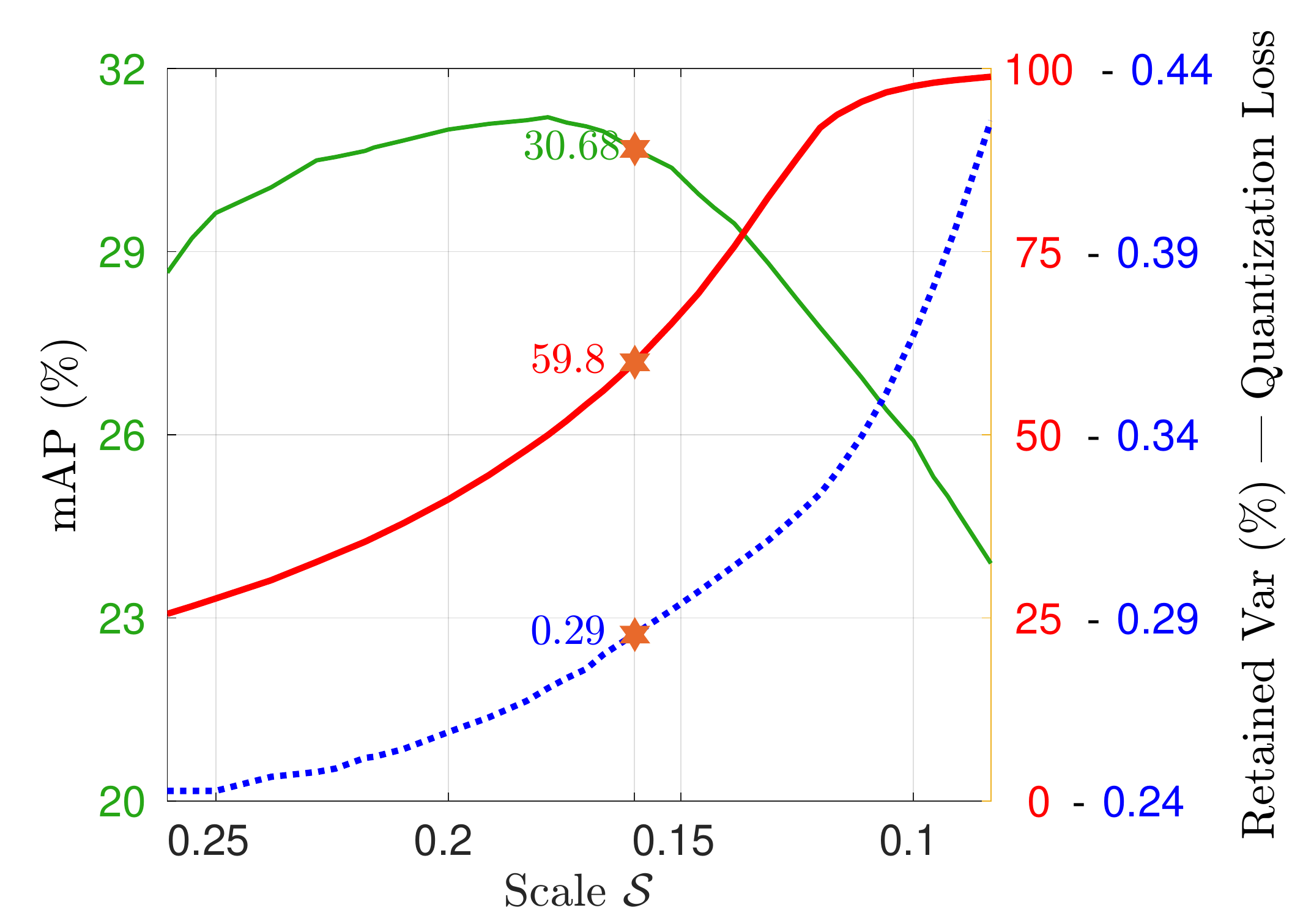}
\caption{CIFAR-10}
\label{fig:cifar10_32}
\end{subfigure}
\begin{subfigure}[b]{0.32\textwidth}
\includegraphics[width=\textwidth]{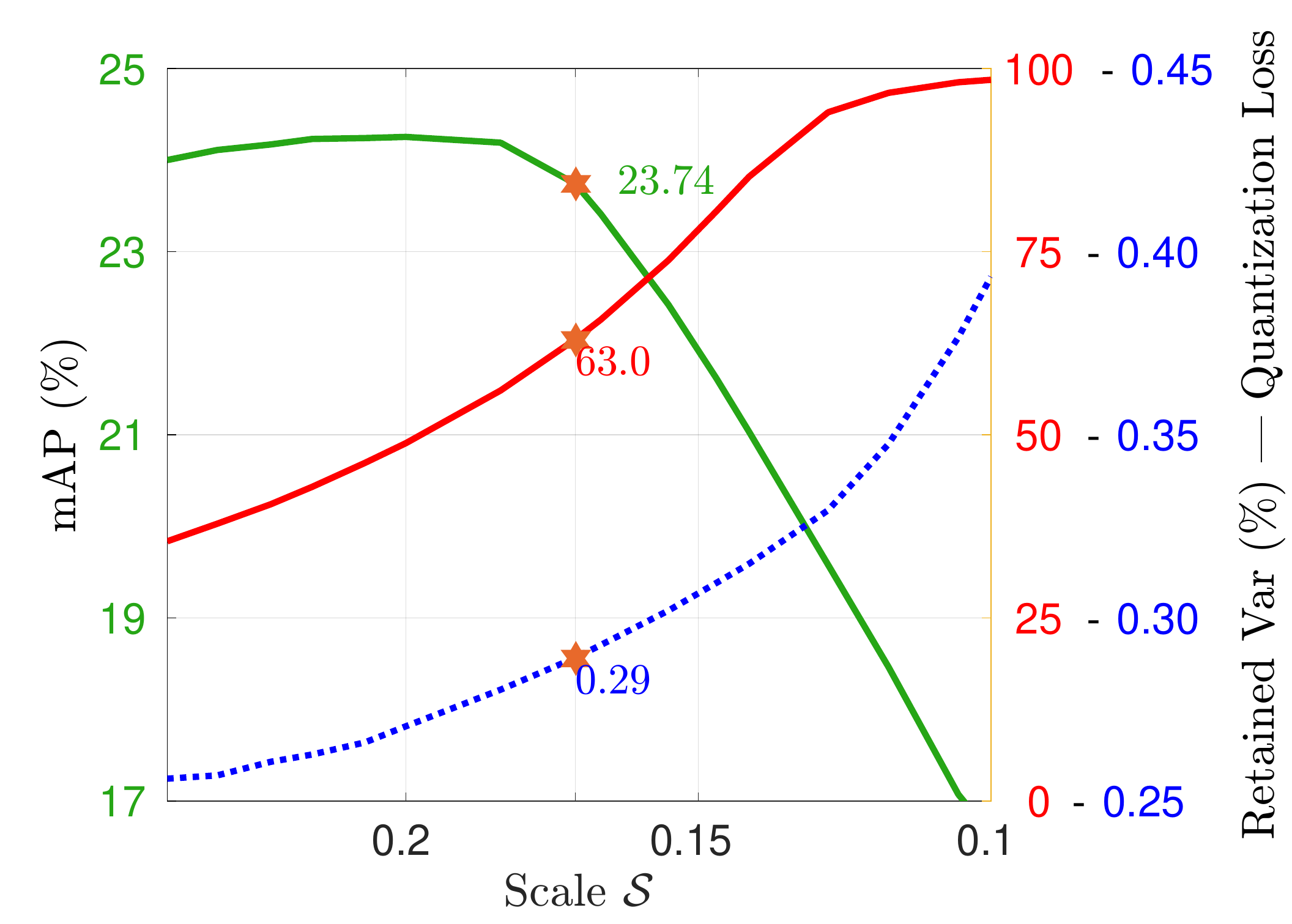}
\caption{Labelme-12-50k}
\label{fig:labelme_32}
\end{subfigure}
\begin{subfigure}[b]{0.32\textwidth}
\includegraphics[width=\textwidth]{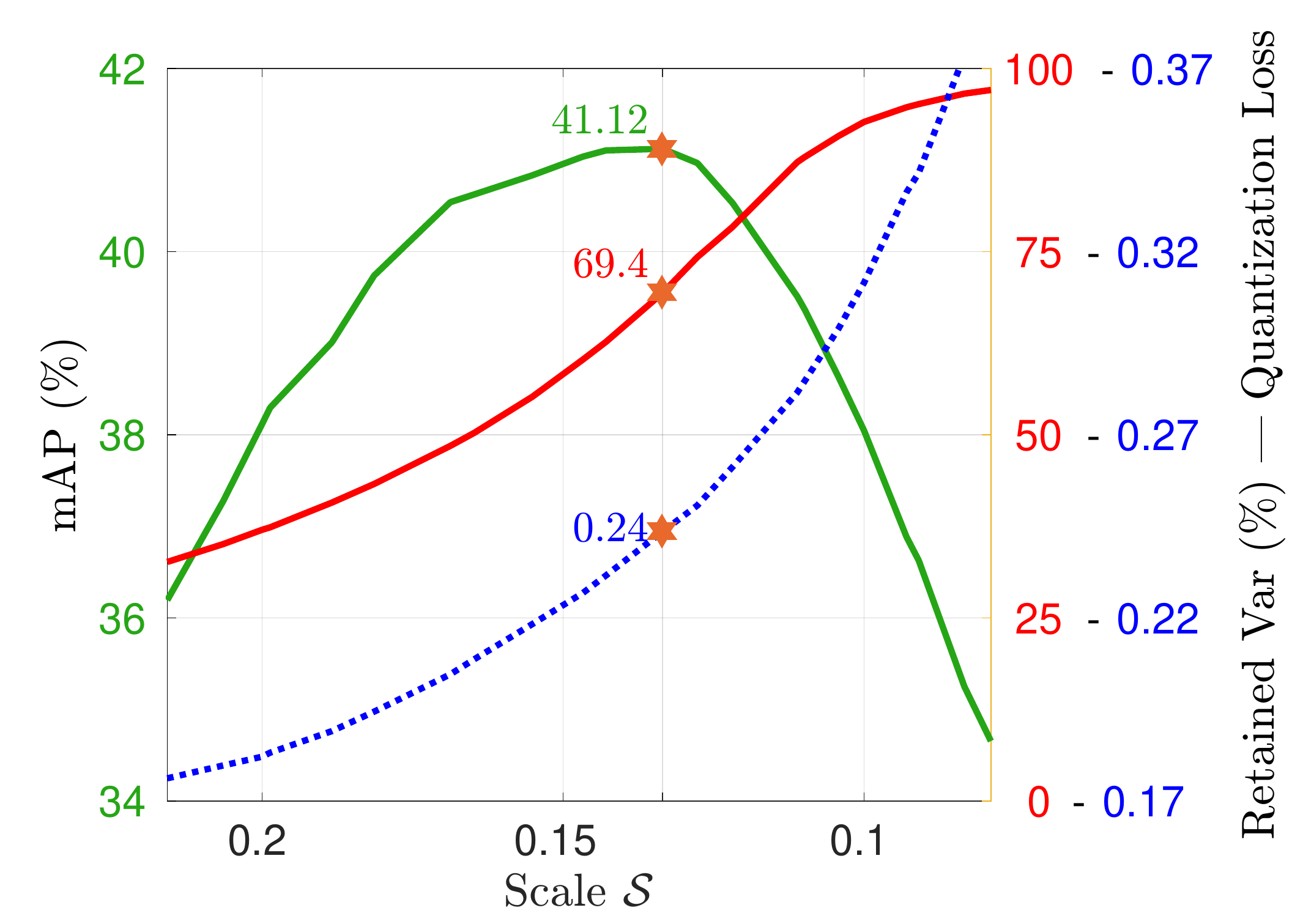}
\caption{SUN397}
\label{fig:sun397_32}
\end{subfigure}
\caption{Analyzing the effects of the scale value $\Sc$ on \textit{(i)} the quantization loss per bit ({\color{black}blue dash line} with {\color{black}blue right Y-axis}), \textit{(ii)} the percentage of total retained variance by the minimizing quantization loss projection matrix (using Algorithm \ref{algo1}) in comparison with the total retained variance of top-$L$ PCA components ({\color{black}red line} with {\color{black}red right  Y-axis}), and \textit{(iii)} the retrieval performance in mAP (green
line with green left Y-axis). 
Note that x-axis is in descending order.}
\label{fig:var_loss_tradeoff}
\end{figure*}

\begin{table*}[t]
\centering
\captionsetup{justification=centering}
\caption{Performance comparison with the state-of-the-art unsupervised hashing methods. The \textbf{Bold} and \underline{Underline} values indicate the \textbf{best} and \underline{second best} performances respectively.}
\label{tb:state-of-art}

\begin{tabular}{|c|l|c|c|c|c|c|c|c|c|c|c|c|c|}
\hline
 & Dataset & \multicolumn{4}{c|}{\textbf{CIFAR-10}} & \multicolumn{4}{c|}{\textbf{LabelMe-12-50k}} & \multicolumn{4}{c|}{\textbf{SUN397}} \\
 \cline{2-14}
& {L} & 8 & 16 & 24 & 32 & 8 & 16 & 24 & 32 & 8 & 16 & 24 & 32 \\
\hline
\hline
\multirow{5}{*}{\rotatebox[origin=c]{90}{\textbf{\textit{mAP}}}}& SpH & 17.09 & 18.77 & 20.19 & 20.96 & 11.68 & 13.24 & 14.39 & 14.97 & 9.13 & 13.53 & 16.63 & 19.07\\
& KMH & 22.22 & 24.17 & 24.71 & 24.99 & 16.09 & 16.18 & 16.99 & 17.24 & 21.91 & 26.42 & 28.99 & 31.87\\
& BA & 23.24 & 24.02 & 24.77 & 25.92 & 17.48 & 17.10 & 17.91 & 18.07 & 20.73 & 31.18 & 35.36 & 36.40\\
& ITQ & 24.75 & 26.47 & 26.86 & 27.19 & 17.56 & 17.73 & 18.52 & 19.09 & 20.16 & 30.95 & 35.92 & 37.84 \\
& \textbf{SCQ - OnE} & \textbf{27.08} & \textbf{29.64} & \underline{30.57} & \underline{30.82} & \underline{19.76} & \underline{21.96} & \textbf{23.61} & \underline{24.25} & \underline{23.37} & \underline{34.09} & \underline{38.13} &  \underline{40.54} \\
& \textbf{SCQ - OgE} & \underline{26.98} & \underline{29.33} & \textbf{30.65} & \textbf{31.15} & \textbf{20.63} & \textbf{23.07} & \underline{23.54} & \textbf{24.68} & \textbf{23.44} & \textbf{34.73} & \textbf{39.47} & \textbf{41.82} \\
\hline
\multirow{5}{*}{\rotatebox[origin=c]{90}{\textbf{\textit{prec@r2}}}}& SpH & 18.04 & 30.58 & 37.28 & 21.40 & 11.72 & 19.38 & 25.14 & 13.66 & 6.88 & 23.68 & 37.21 & 27.39\\
& KMH & 21.97 & 36.64 & 42.33 & 27.46 & 15.20 & 26.17 & 32.09 & 18.62 & 9.50 & \underline{36.14} & \textbf{51.27} & 39.29\\
& BA & 23.67 & 38.05 & 42.95 & 23.49 & 16.22 & 25.75 & 31.35 & 13.14 & \textbf{10.50} & \textbf{37.75} & \underline{50.38} & 41.11\\
& ITQ & \underline{24.38} & \textbf{38.41} & \underline{42.96} & 28.63 & 15.86 & 25.46 & 31.43 & 17.66 & \underline{9.78} & 35.15 & 49.85 & 46.34 \\
& \textbf{SCQ - OnE} & \textbf{24.48} & {36.49} & {41.53} & \underline{43.90} & \textbf{16.69} & \underline{27.30} & \underline{34.63} & \underline{33.04} & {8.68} & {30.12} & 43.54 & \underline{50.41} \\
& \textbf{SCQ - OgE} & {24.35} & \underline{38.30} & \textbf{43.01} & \textbf{44.01} & \underline{16.57} & \textbf{27.80} & \textbf{34.77} & \textbf{34.64} & 8.76 & 29.31 & 45.03 & \textbf{51.88} \\
\hline
\multirow{5}{*}{\rotatebox[origin=c]{90}{\textbf{\textit{prec@1k}}}}& SpH & 22.93 & 26.99 & 29.50 & 31.98 & 14.07 & 16.78 & 18.52 & 19.27 & 10.79 & 15.36 & 18.21 & 20.07\\
& KMH & 32.30 & 33.65 & 35.52 & 37.77 & 21.07 & 20.97 & 21.41 & 21.98 & 18.94 & 24.93 & 25.74 & 28.26\\
& BA  & 31.73 & 34.16 & 35.67 & 37.01 & 21.14 & 21.71 & 22.64 & 22.83 & 19.22 & 28.68 & 31.31 & 31.80 \\
& ITQ & 32.40 & 36.35 & 37.25 & 37.96 & 21.01 & 22.00 & 22.98 & 23.63 & 18.86 & 28.62 & 31.56 & 32.74 \\
& \textbf{SCQ - OnE} & \underline{33.38} & \underline{37.82} & \underline{39.13} & \underline{40.40} & \underline{22.91} & \underline{25.39} & \underline{26.55} & \underline{27.16} & \underline{19.26} & \underline{29.95} & \underline{32.72} & \underline{34.08} \\
& \textbf{SCQ - OgE} & \textbf{33.41} & \textbf{38.33} & \textbf{39.54} & \textbf{40.70} & \textbf{23.94} & \textbf{25.94} & \textbf{26.99} & \textbf{27.46} & \textbf{20.10} & \textbf{29.95} & \textbf{33.43} & \textbf{35.00} \\
\hline
\end{tabular}
\end{table*}

The \textbf{CIFAR-10} dataset (\cite{cifar10}) contains $60,000$ fully-annotated color images of $32\times 32$ from $10$ object classes ($6,000$ images for each class). The provided test set ($1,000$ images for each class) is used as the query set. The remaining 50,000 images are used as the training set and the database. 

The \textbf{LabelMe-12-50k} dataset (\cite{labelme_12_50k}) consists of $50,000$ fully annotated color images of $256 \times 256$ of $12$ object classes, which is a subset of LabelMe dataset (\cite{labelme_full}). In this dataset, for any image having multiple label values in the range of $[0.0, 1.0]$, the object class of the largest label value is chosen as the image label. We also use the provided test set as the query set and the remaining images as the training set and the database.

The \textbf{SUN397} dataset (\cite{SUN397}) contains approximately $108,000$ fully annotated color images from $397$ scene categories. 
We select a subset of $42$ categories which contain more than 500 images per category to construct a dataset of approximately $35,000$ images in total. We then randomly sample 100 images per class to form the query set. The remaining images are used as the training set and the database.

For these above image datasets, each image is represented by a 4096-D feature vector extracted from the fully-connected layer 7 of pre-trained VGG (\cite{VGG}). 



\textbf{Evaluation protocols.} 
As datasets are fully annotated, we use semantic labels to define the ground truths of image queries. 
We apply three standard evaluation metrics, which are widely used in literature (\cite{BA,DeepHash, ITQ}), to measure the retrieval performance of all methods: 
1) mean Average Precision (\textbf{\textit{mAP}}($\%$)); 
2) precision at Hamming radius of 2 (\textbf{\textit{prec@r2}} ($\%$)) which measures precision on retrieved images having Hamming distance to query $\le 2$ (we report zero precision for queries that return no image); 
3) precision at top 1000 return images (\textbf{\textit{prec@1k}} ($\%$)) which measures the precision on the top 1000 retrieved images. 


\textbf{Implementation notes.}
As discussed in section \ref{sssec:complexity}, for computational efficiency, we apply PCA to reduce the feature dimension to 512-D for our proposed methods. 
The hyper-parameter $\mu$ of OgE algorithm is empirically set as $0.02$ for all experiments. Finally, for both OnE and OgE, we set all error-tolerance values, $\epsilon, \epsilon_b, \epsilon_n$, as $10^{-4}$ and the maximum number of iteration is set as $100$. 
The implementation of our methods is available at \url{https://github.com/hnanhtuan/SCQ.git}.

For all compared methods, e.g., Spherical Hashing (SpH) (\cite{SpH}), K-means Hashing (KMH)\footnote{Due to very long training time at high-dimension of KMH (\cite{KMH}), we apply PCA to reduce dimension from 4096-D to 512-D. Additionally, we execute experiments for KMH with $b=\{2,4,8\}$ and report the best results.} (\cite{KMH}), Binary Autoencoder (BA) (\cite{BA}), and Iterative Quantization (ITQ) (\cite{ITQ}); we use the implementation with suggested parameters provided by the authors. Besides, to improve the statistical stability of the results, we report the average values of 5 executions.


\subsection{Effects of parameters}
\label{ssec:effect_of_param}

As discussed in section \ref{ssec:var_vs_quan}, when $\Sc$ decreases, the projection matrix $\V$ can be learned to retain a very high amount of variance, as much as PCA can. However, it causes undesirable large binary quantization loss and vice versa. 
In this section, we additionally provide quantitative analysis of the effects of the scale parameter on these two factors (i.e., the amount of retained variance and the quantization loss) and, moreover, on the retrieval performance.



In this experiment, for all datasets, e.g., CIFAR-10, LabelMe-12-50k, and SUN397, we random select 20 images for each class in the training set (as discussed in section \ref{ssec:datasets}) for validation set. The remaining images are used for training.
To obtain each data point, we solve the problem Eq. (\ref{eq:OnE}) at various scale values $\Sc$
and use OnE algorithm (Algorithm \ref{algo1} - Section \ref{ssec:OnE_opt}) to tackle the optimization. 


Figure \ref{fig:var_loss_tradeoff} presents (i) the quantization loss per bit, (ii) the percentage of total retained variance by the minimizing quantization loss projection matrix in comparison with the total retained variance of top-$L$ PCA components as $\Sc$ varies, and (iii) the retrieval performance (\textbf{\textit{mAP}}) of the validation sets.
Firstly, we can observe that there is no scale $\Sc$ that can simultaneously maximizes the retained variance and minimizes the optimal quantization loss. On the one hand, as the scale value $\Sc$ decreases, minimizing the loss function Eq. (\ref{eq:rewriteOnE}) produces a projection matrix that focuses on high-variance directions, i.e., retains more variance in comparison with PCA (red line). On the other hand, at smaller $\Sc$, the quantization loss is much larger (blue dash line). The empirical results are consistent with our discussion in section \ref{ssec:var_vs_quan}.

Secondly, regarding the retrieval performance, unsurprisingly, the performance drops as the scale $\Sc$ gets too small, i.e., a high amount of variance is retained but the quantization loss is too large, or $\Sc$ gets too large, i.e., the quantization loss is small but only low variance components are retained.  
Hence, it is necessary to balance these two factors. As data variance varies from dataset to dataset, the scale value should be determined from the dataset. In particular, we leverage the eigenvalues $\Lambda$, which are the variances of PCA components, to determine this hyper-parameter. 
From experimental results in Figure \ref{fig:var_loss_tradeoff}, we propose to formulate the scale parameter as:
\begin{equation}
\Sc = \sqrt[]{\frac{L}{\sum_{i=1}^L \lambda_i}},
\label{eq:scale}
\end{equation}
One advantage of this setting is that it can generally achieve the best performances across multiple datasets, feature types, and hash lengths, without resort to conducting multiple trainings and cross-validations. The proposed working points of the scale are shown in Figure \ref{fig:var_loss_tradeoff}. We apply this scale parameter to the datasets for both OnE and OgE algorithms in all later experiments. 

Note that the numerator of the fraction in Eq. \ref{eq:scale}, i.e., $L$ is the hash code length, which is also the total variance of binary codes $\B$. 
In addition, the denominator is the total variance of top $L$-th PCA components, i.e., the maximum amount of variance that can be retained in an $L$-dimension feature space. Hence, we can interpret the scale as a factor that make the amounts of variance, i.e., energy, of the input $\X$ and output (i.e. binary codes $\B$) are comparable. This property is important as when the variance of input is much larger than the variance of output, obviously there is some information loss. On the other hand, when the variance of output is larger than it of input, the output contains undesirable additional information. 

\red{
\begin{table}[t]
\small
\centering
\caption{Summary of the percentage of retained variance (\%), quantization loss per bit, and retrieval performance (\textbf{\textit{mAP}}) on validation sets for ITQ and our SCQ-OnE methods (at the proposed scale of Eq. \eqref{eq:scale}).}
\label{tb:summary}
\begin{tabular}{|c|c|c|c|c|}
\hline
 & Method & CIFAR-10 & LabelMe &  SUN397  \\ \hline
\% Retained & ITQ & 100\% & 100\% & 100\% \\ \cline{2-5}
 variance & SCQ-OnE & 59.6\% & 63.0\% & 69.4\% \\
\hline
Quantization & ITQ & 0.75 & 0.71 & 0.65 \\\cline{2-5}
 error & SCQ-OnE & 0.29 & 0.29 & 0.24 \\
 \hline
\multirow{2}{*}{\textbf{\textit{mAP}}} & ITQ & 27.01 & 18.24 & 37.79 \\\cline{2-5}
 & SCQ-OnE & \textbf{30.68} & \textbf{23.74} & \textbf{41.12} \\
 \hline
\end{tabular}
\end{table}

Additionally, in Table \ref{tb:summary}, we summarize the percentage of retained variance (\%), quantization loss per bit, and retrieval performance (\textbf{\textit{mAP}}) on validation sets for ITQ and our SCQ-OnE methods. Even though, the projection matrix, learned by our Algorithm 1, can retain less variance in comparison to the optimal PCA projection matrix (i.e., the ITQ first step), this helps to achieve a much smaller quantization error. Hence, balancing the variance loss and quantization error is desirable and can result in higher retrieval performance.
}

\subsection{Comparison with state-of-the-art}

In this section, we evaluate our proposed hashing methods, SCQ - OnE and OgE, and compare to the state-of-the-art unsupervised hashing methods including SpH, KMH, BA, and ITQ.
The experimental results in \textbf{\textit{mAP}}, \textbf{\textit{prec@r2}} and \textbf{\textit{prec@1k}} are reported in Table \ref{tb:state-of-art}. Our proposed methods clearly achieve significant improvement over all datasets at the majority of evaluation metrics.  The improvement gaps are clearer at higher code lengths,
i.e., $L = 32$. Additionally, OgE generally achieves slightly higher performance than OnE. 
Moreover, it is noticeable that, for \textbf{\textit{prec@r2}}, all compared methods suffer performance downgrade at long hash code, e.g., $L=32$. However, our proposed methods still achieve good \textbf{\textit{prec@r2}} at $L=32$. This shows that binary codes producing by our methods highly preserve data similarity.

\smallskip

\begin{table}[t]
\small
\caption{Performance comparison in \textbf{\textit{mAP}} and \textbf{\textit{prec@r2}} with Deep Hashing (DH) (\cite{DeepHash}) and Unsupervised Hashing with Binary Deep Neural Network (UH-BDNN) (\cite{UH-BDNN}) on CIFAR-10 dataset for $L=16$ and $32$. The \textbf{Bold} values indicate the \textbf{best} performances.}
\label{tb:DH-UH_BDNN}
\centering
\begin{tabular}{|c|l|c|c|c|c|}
\hline
& \multirow{2}{*}{Methods}
& \multicolumn{2}{c|}{\textbf{\textit{mAP}}} & \multicolumn{2}{c|}{\textbf{\textit{prec@r2}}} \\ 
\cline{3-6}
& & 16 & 32 & 16 & 32 \\ 
\hline
\multirow{4}{*}{\rotatebox[origin=c]{90}{\textbf{CIFAR-10}}}
& DH   & 16.17 & 16.62 & 23.33 & 15.77 \\
& UH-BDNN  & 17.83 &18.52& \textbf{24.97}& 18.85 \\

& {SCQ - OnE} & {17.97} & {18.63} & 24.57 & {23.72} \\
& {SCQ - OgE} & \textbf{18.00} & \textbf{18.78} & 24.15 & \textbf{25.69} \\
\hline
\end{tabular}
\end{table}

\begin{table*}[t]
\small
\centering
\caption{Performance comparison in \textbf{\textit{mAP}} with BGAN (\cite{BGAN}) on CIFAR-10 and NUS-WIDE datasets.}
\label{tb:compare_BGAN}
\begin{tabular}{|c|l|c|c|c|c|c|c|c|c|}
\hline
& \multirow{2}{*}{Methods} & \multicolumn{4}{c|}{\textbf{CIFAR-10}} & \multicolumn{4}{c|}{\textbf{NUS-WIDE}}\\
\cline{3-10}
& & 12 & 24 & 32 & 48 & 12 & 24 & 32 & 48 \\
\hline
\multirow{3}{*}{\rotatebox[origin=c]{90}{\textbf{\textit{mAP}}}} & BGAN  & 40.1 & 51.2 & 53.1 & 55.8 & 67.5 & 69.0 & 71.4 & 72.8 \\
& SCQ - OnE & 53.59 & \textbf{55.77} & 57.62 & 58.14 & 69.82 & 70.53 & \textbf{72.78} & \textbf{73.25} \\
& SCQ - OgE & \textbf{53.83} & 55.65 & \textbf{57.74} & \textbf{58.44} & \textbf{70.17} & \textbf{71.31} & 72.49 & 72.95 \\
\hline
\end{tabular}
\end{table*}

\textbf{Comparison with Deep Hashing (DH) (\cite{DeepHash}) and Unsupervised Hashing with Binary Deep Neural Network (UH-BDNN) (\cite{UH-BDNN}).}
Recently, there are several methods (\cite{DeepHash,UH-BDNN}) applying DNN to learn binary hash codes. These method can achieve very competitive performances. Hence, in order to have a complete evaluation, following the experiment settings of \cite{DeepHash,UH-BDNN}, we conduct experiments on the CIFAR-10 dataset. In this experiment, 100 images are randomly sampled for each class as a query set; the remaining images are for training and database. Each image is presented by a GIST 512-D descriptor (\cite{GIST}). In addition, to avoid bias results due to test samples, we repeat the experiment 5 times with 5 different random training/query sets. 
The comparative results in term of \textbf{\textit{mAP}} and \textbf{\textit{prec@r2}} are presented in Table \ref{tb:DH-UH_BDNN}. Our proposed methods are very competitive with DH and UH-BDNN, specifically achieving higher \textbf{\textit{mAP}} and \textbf{\textit{prec@r2}} at $L=32$ than DH and UH-BDNN.

\textbf{Comparison with Binary Generative Adversarial Networks for Image Retrieval (BGAN) (\cite{BGAN}).} Recently, BGAN applies a continuous approximation of \textit{sign} function to learn the binary codes which can help to generate images plausibly similar to the original images. The method has been proven to achieve outstanding performances in unsupervised image hashing task. It is important to note that BGAN is different from our method and compared methods in the aspect that BGAN jointly learns image feature representations and binary codes, in which the binary codes are achieved by using an approximate smooth function of \textit{sign}.
While ours and compared methods learn the optimal binary codes given image representations. 
Hence, to further validate the effectiveness of our methods and to compare with BGAN, we apply our method on the FC7 features extracted from the feature extraction component in the pre-trained BGAN model\footnote{The model is obtained after training BGAN method on CIFAR-10 and NUS-WIDE datasets accordingly. The same model is also used to obtain BGAN binary codes.} on CIFAR-10 and NUS-WIDE (\cite{nuswide}) datasets. 
In this experiment, we aim to show that by applying our hashing methods on the pretrained features from feature extraction component of BGAN, 
our methods can produce better hash codes than the hash codes which are obtained from the jointly learning approach of BGAN.

Similar to the experiment setting in BGAN (\cite{BGAN}), for both CIFAR-10 and NUS-WIDE, we randomly select 100 images per class as the test query set; the remaining images are used as database for retrieval. We then randomly sample from the database set 1,000 images per class as the training set. 
The Table \ref{tb:compare_BGAN} shows that by using 
the more discriminative features\footnote{In comparison with the image features which are obtained from the pre-trained off-the-shelf VGG network (\cite{VGG}).} from the pre-trained feature extraction component of BGAN, our methods can outperform BGAN, i.e., our methods can produce better binary codes in comparison to the \textit{sign} approximate function in BGAN, and achieve the state-of-the-art performances in the unsupervised image hashing task. Hence, the experiment results emphasize the important of an effective method to preserve the discrimination power of high-dimensional CNN features in very compact binary representations, i.e., effectively handling the challenging binary and orthogonal constraints.


\subsection{Training time and Processing time}
\label{ssec:train_time}

\begin{figure}[t]
\centering
\includegraphics[width=0.33\textwidth]{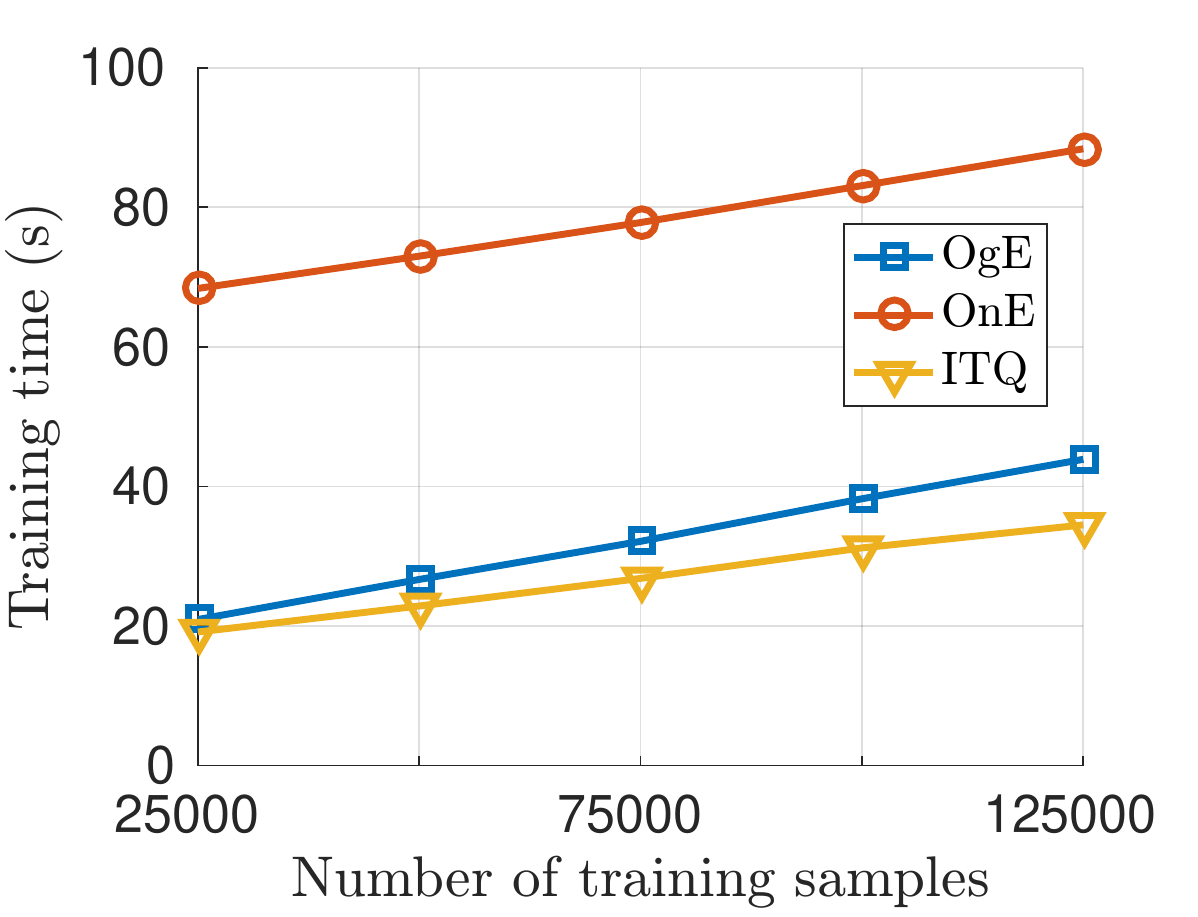}
\caption{The training time for learning 32-bit hash code embedding.}
\label{fig:training_time}
\end{figure}

In this experiment, we empirically evaluate the training time and online processing time of our methods. The experiments are carried out on a workstation with a 4-core i7-6700 CPU @ 3.40GHz. The experiments are conducted on the combination of CIFAR-10, Labelme-12-50k, and SUN397 datasets. 
For OnE and OgE, the training time include time for applying zero-mean, scaling, reducing dimension to $D=512$. We use 50 iterations for all experiments. The Fig. \ref{fig:training_time} shows that our proposed methods, OnE and OgE, are very efficient. OgE is just slightly slower than ITQ. Even though OnE is slower than OgE and ITQ, it takes just over a minute for 100.000 training samples which is still very fast and practical, in comparison with several dozen minutes for KMH, BA, and UH-BDNN\footnote{For training 50000 CIFAR-10 samples using author's release code and dataset (\cite{UH-BDNN}).}.

Compared with training cost, the time to produce new hash codes is more important since it is done in real time. Similar to Semi-Supervised Hashing (SSH) (\cite{SemiSupHash}) and ITQ (\cite{ITQ}), by using only a single linear transformation, our proposed methods require only one BLAS operation ($\texttt{gemv}$ or $\texttt{gemm}$) and a comparison operation; hence, it takes negligible time to produce binary codes for new data points.

\section{Conclusion}
\label{sec:conclusion}
In this paper, we successfully addressed the problem of jointly learning to preserve data pairwise (dis)similarity in low-dimension space and to minimize the binary quantization loss with the strict diagonal constraint. 
Additionally, we show that as more variance is retained, the quantization loss is undesirably larger; and vice versa. Hence, by appropriately balancing these two factors using a scale, our methods can produce better binary codes.
Extensive experiments on various datasets show that our proposed methods, Simultaneous Compression and Quantization (SCQ): Orthonormal Encoder (OnE) and Orthogonal Encoder (OgE), outperform  other state-of-the-art hashing methods by clear margins under various standard evaluation metrics and benchmark datasets. 
Furthermore, OnE and OgE are very computationally efficient in both training and testing steps.

\bibliographystyle{model2-names}
\bibliography{refs}
\begin{appendices}
\red{
\section{Derivation for Eq. \eqref{eq:nu_cond}}
\label{sec:derivation_1}

Firstly, the dual function $\mathcal{G}_1(\nu_1)$ can be simply constructed by substituting $\bV_1$ from Eq. (7) into Eq. (6):
\begin{equation}
\begin{split}
    \mathcal{G}(\nu_1) =& \frac{1}{n}\left\|\bb_1 - \bX(\bX^\top\bX + n\nu_1 \mathbf{I})^{-1}\bX^\top\bb_1\right\|^2 \\ +\nu_1&\left(\bb_1^\top\bX(\bX^\top\bX + n\nu_1 \mathbf{I})^{-\top}(\bX^\top\bX + n\nu_1 \mathbf{I})^{-1}\bX^\top\bb_1 - 1\right)
\end{split}
\end{equation}

Firstly, we note that:
{\small
\begin{equation}
\begin{split}
    &\frac{\partial (\bX^\top\bX + n\nu_1 \mathbf{I})^{-1}}{\partial \nu_1}\\
    &= -(\bX^\top\bX + n\nu_1 \mathbf{I})^{-1}\frac{\partial (\bX^\top\bX + n\nu_1 \mathbf{I})}{\partial \nu_1}(\bX^\top\bX + n\nu_1 \mathbf{I})^{-1}\\
    &=-n(\bX^\top\bX + n\nu_1 \mathbf{I})^{-1}(\bX^\top\bX + n\nu_1 \mathbf{I})^{-1}
\end{split}
\end{equation}
}

Hence,
{\small
\begin{equation}
\begin{split}
&\frac{\partial \mathcal{G}(\nu_1)}{\partial \nu_1}\\
=&- \frac{2}{n}\left(\bb_1 - \bX(\bX^\top\bX + n\nu_1 \mathbf{I})^{-1}\bX^\top\bb_1\right)^\top\bX\frac{\partial (\bX^\top\bX + n\nu_1 \mathbf{I})^{-1}}{\partial \nu_1}\bX^\top\bb_1\\
&+\left(\bb_1^\top\bX(\bX^\top\bX + n\nu_1 \mathbf{I})^{-\top}(\bX^\top\bX + n\nu_1 \mathbf{I})^{-1}\bX^\top\bb_1 - 1\right)\\
&+ 2\nu_1\bb_1^\top\bX(\bX^\top\bX + n\nu_1 \mathbf{I})^{-\top}\frac{\partial (\bX^\top\bX + n\nu_1 \mathbf{I})^{-1}}{\partial \nu_1}\bX^\top\bb_1\\
=&2\bb_1^\top\bX(\bX^\top\bX + n\nu_1 \mathbf{I})^{-\top}(\bX^\top\bX + n\nu_1 \mathbf{I})^{-1}\bX^\top\bb_1\\
-&2\bb_1^\top\bX(\bX^\top\bX + n\nu_1 \mathbf{I})^{-\top}\bX^\top\bX(\bX^\top\bX + n\nu_1 \mathbf{I})^{-1}(\bX^\top\bX + n\nu_1 \mathbf{I})^{-1}\bX^\top\bb_1\\
+&\left(\bb_1^\top\bX(\bX^\top\bX + n\nu_1 \mathbf{I})^{-\top}(\bX^\top\bX + n\nu_1 \mathbf{I})^{-1}\bX^\top\bb_1 - 1\right)\\
-&2n\nu_1\bb_1^\top\bX(\bX^\top\bX + n\nu_1 \mathbf{I})^{-\top}(\bX^\top\bX + n\nu_1 \mathbf{I})^{-1}(\bX^\top\bX + n\nu_1 \mathbf{I})^{-1}\bX^\top\bb_1\\
=&2\bb_1^\top\bX(\bX^\top\bX + n\nu_1 \mathbf{I})^{-\top}(\bX^\top\bX + n\nu_1 \mathbf{I})^{-1}\bX^\top\bb_1\\
-&2\bb_1^\top\bX(\bX^\top\bX + n\nu_1 \mathbf{I})^{-\top}(\bX^\top\bX+n\nu_1\bI)(\bX^\top\bX + n\nu_1 \mathbf{I})^{-1}(\bX^\top\bX + n\nu_1 \mathbf{I})^{-1}\bX^\top\bb_1\\
+&2n\nu_1\bb_1^\top\bX(\bX^\top\bX + n\nu_1 \mathbf{I})^{-\top}(\bX^\top\bX + n\nu_1 \mathbf{I})^{-1}(\bX^\top\bX + n\nu_1 \mathbf{I})^{-1}\bX^\top\bb_1\\
+&\left(\bb_1^\top\bX(\bX^\top\bX + n\nu_1 \mathbf{I})^{-\top}(\bX^\top\bX + n\nu_1 \mathbf{I})^{-1}\bX^\top\bb_1 - 1\right)\\
-&2n\nu_1\bb_1^\top\bX(\bX^\top\bX + n\nu_1 \mathbf{I})^{-\top}(\bX^\top\bX + n\nu_1 \mathbf{I})^{-1}(\bX^\top\bX + n\nu_1 \mathbf{I})^{-1}\bX^\top\bb_1\\
=&\bb_1^\top\bX(\bX^\top\bX + n\nu_1 \mathbf{I})^{-\top}(\bX^\top\bX + n\nu_1 \mathbf{I})^{-1}\bX^\top\bb_1 - 1\\
\end{split}
\end{equation}

\begin{equation}
\label{eq:2nd_cond_1}
\Leftrightarrow \frac{\partial \mathcal{G}_1}{\partial \nu_1} = \bb_1^\top\bX(\bX^\top\bX + n\nu_1 \mathbf{I})^{-\top}(\bX^\top\bX + n\nu_1 \mathbf{I})^{-1}\bX^\top\bb_1 - 1
\end{equation}
}
Note that: $(\bX^\top\bX + n\nu_1 \mathbf{I})^{-1} = (\bX^\top\bX + n\nu_1 \mathbf{I})^{-\top}$.
\\

A simpler way to achieve $\frac{\partial \mathcal{G}(\nu_1)}{\partial \nu_1}$ is to take the deravative of $\mathcal{L}$ w.r.t $\nu_1$ first, then replace $\bv_1$ by Eq. (7) later.

\begin{equation}
    \frac{\partial\mathcal{L}}{\partial \nu_1} = \bv_1^\top\bv_1-1
\end{equation}
\begin{equation}
\label{eq:2nd_cond_2}
\Rightarrow \frac{\partial \mathcal{G}}{\partial \nu_1} = \bb_1^\top\bX(\bX^\top\bX + n\nu_1 \mathbf{I})^{-\top}(\bX^\top\bX + n\nu_1 \mathbf{I})^{-1}\bX^\top\bb_1 - 1\\
\end{equation}
By setting $\frac{\partial\mathcal{G}_1}{\partial \nu_1}=0$ (Eq. \eqref{eq:2nd_cond_2}), we can obtain the second condition in Eq. \eqref{eq:nu_cond}.

\section{Derivation for Eq. \eqref{eq:nu_k_cond}}
\label{sec:derivation_2}
Following the similar derivation in Appendix section \ref{sec:derivation_1}, we can obtain the second condition of Eq. \eqref{eq:nu_k_cond}. We now provide the detail derivation for the third condition. Considering the $i$-th value ($\phi_{ki}$) of the Lagrange multiplier $\Phi_k$
\begin{equation}
\frac{\partial\mathcal{L}_k}{\partial\phi_{ki}} = \bv_k^\top\bv_i
\end{equation}

{\small
\begin{equation}
\begin{split}
\Rightarrow \frac{\partial\mathcal{G}_k}{\partial\phi_{ki}} &= \left(\mathbf{X}^\top\mathbf{b}_k-\frac{n}{2}\sum_{j=1}^{k-1}\phi_{kj}\mathbf{v}_j\right)^\top (\mathbf{X}^\top\mathbf{X}+n\nu_k \mathbf{I})^{-\top}\bv_i\\
&=-\frac{n}{2}\phi_{k1}\bv_1^\top\bZ_k\bv_i - \cdots - \frac{n}{2}\phi_{k(k-1)}\bv_{(k-1)}^\top\bZ_k\bv_1 + \bb_k^\top\bX\bZ_k\bv_i
\end{split}
\end{equation}
}
where $\bZ_k = (\mathbf{X}^\top\mathbf{X}+n\nu_k \mathbf{I})^{-1}$.

By setting the derivative of $\mathcal{G}_k$ w.r.t $\Phi_k= [\phi_{k1}, ..., \phi_{k(k-1)}]^\top$ equal to $[0, \cdots, 0]$ and some simple manipulations, we can obtain the third condition of Eq. \eqref{eq:nu_k_cond} as follows:
\begin{equation}
    \mathbf{A}_k\Phi_k = \mathbf{c}_k,
\end{equation}
where
\begin{equation}
\begin{cases}
\mathbf{A}_k = \frac{n}{2}\begin{bmatrix} \mathbf{v}_1^\top\mathbf{Z}_k\mathbf{v}_1 & \cdots & \mathbf{v}_1^\top\mathbf{Z}_k\mathbf{v}_{(k-1)} \\
\vdots & \ddots & \vdots \\
\mathbf{v}_{(k-1)}^\top\mathbf{Z}_k\mathbf{v}_1 & \cdots &  \mathbf{v}_{(k-1)}^\top\mathbf{Z}_k\mathbf{v}_{(k-1)}
\end{bmatrix}\\
\mathbf{c}_k = \begin{bmatrix}
\mathbf{v}_1^\top\mathbf{Z}_k\mathbf{X}^\top\mathbf{b}_k&
\cdots &
\mathbf{v}_{(k-1)}^\top\mathbf{Z}_k\mathbf{X}^\top\mathbf{b}_k
\end{bmatrix}^\top
\end{cases}
\end{equation}
}

\end{appendices}


\end{document}